\documentclass{article}

\usepackage{microtype}
\usepackage{graphicx}
\usepackage{subfigure}
\usepackage{booktabs} 

\usepackage{hyperref}

\usepackage[accepted]{icml2024}

\usepackage{amsmath}
\usepackage{amssymb}
\usepackage{mathtools}
\usepackage{amsthm}
\usepackage{float}

\usepackage[capitalize,noabbrev]{cleveref}

\theoremstyle{plain}
\newtheorem{theorem}{Theorem}[section]
\newtheorem{proposition}[theorem]{Proposition}

\theoremstyle{definition}

\theoremstyle{remark}

\usepackage[textsize=tiny]{todonotes}

\begin{document}

\twocolumn[
\icmltitle{DIDA: Denoised Imitation Learning based on Domain Adaptation}



\icmlsetsymbol{equal}{*}

\begin{icmlauthorlist}
\icmlauthor{Kaichen Huang}{equal,yyy,comp}
\icmlauthor{Hai-Hang Sun}{equal,yyy,comp}
\icmlauthor{Shenghua Wan}{yyy,comp}
\icmlauthor{Minghao Shao}{yyy,comp}
\icmlauthor{Shuai Feng}{sch}
\icmlauthor{Le Gan}{yyy,comp}
\icmlauthor{De-Chuan Zhan}{yyy,comp}
\end{icmlauthorlist}

\icmlaffiliation{yyy}{National Key Laboratory for Novel Software Technology, Nanjing University, China
}
\icmlaffiliation{comp}{School of Artificial Intelligence, Nanjing University, China}
\icmlaffiliation{sch}{School of ZZZ, Institute of WWW, Location, Country}
\icmlcorrespondingauthor{Kaichen Huang}{huangkc@lamda.nju.edu.cn}

\icmlkeywords{Machine Learning, ICML}

\vskip 0.3in
]




\begin{abstract}
Imitating skills from low-quality datasets, such as sub-optimal demonstrations and observations with distractors, is common in real-world applications. In this work, we focus on the problem of \textit{Learning from Noisy Demonstrations} (LND), where the imitator is required to learn from data with noise that often occurs during the processes of data collection or transmission. Previous IL methods improve the robustness of learned policies by injecting an adversarially learned Gaussian noise into pure expert data or utilizing additional ranking information, but they may fail in the LND setting. To alleviate the above problems, we propose \textbf{D}enoised \textbf{I}mitation learning based on \textbf{D}omain \textbf{A}daptation (\textbf{DIDA}), which designs two discriminators to distinguish the noise level and expertise level of data, facilitating a feature encoder to learn task-related but domain-agnostic representations. Experiment results on MuJoCo demonstrate that DIDA can successfully handle challenging imitation tasks from demonstrations with various types of noise, outperforming most baseline methods.
\end{abstract}

\section{Introduction}

\begin{figure}[ht]
\centering
\includegraphics[width=\columnwidth]{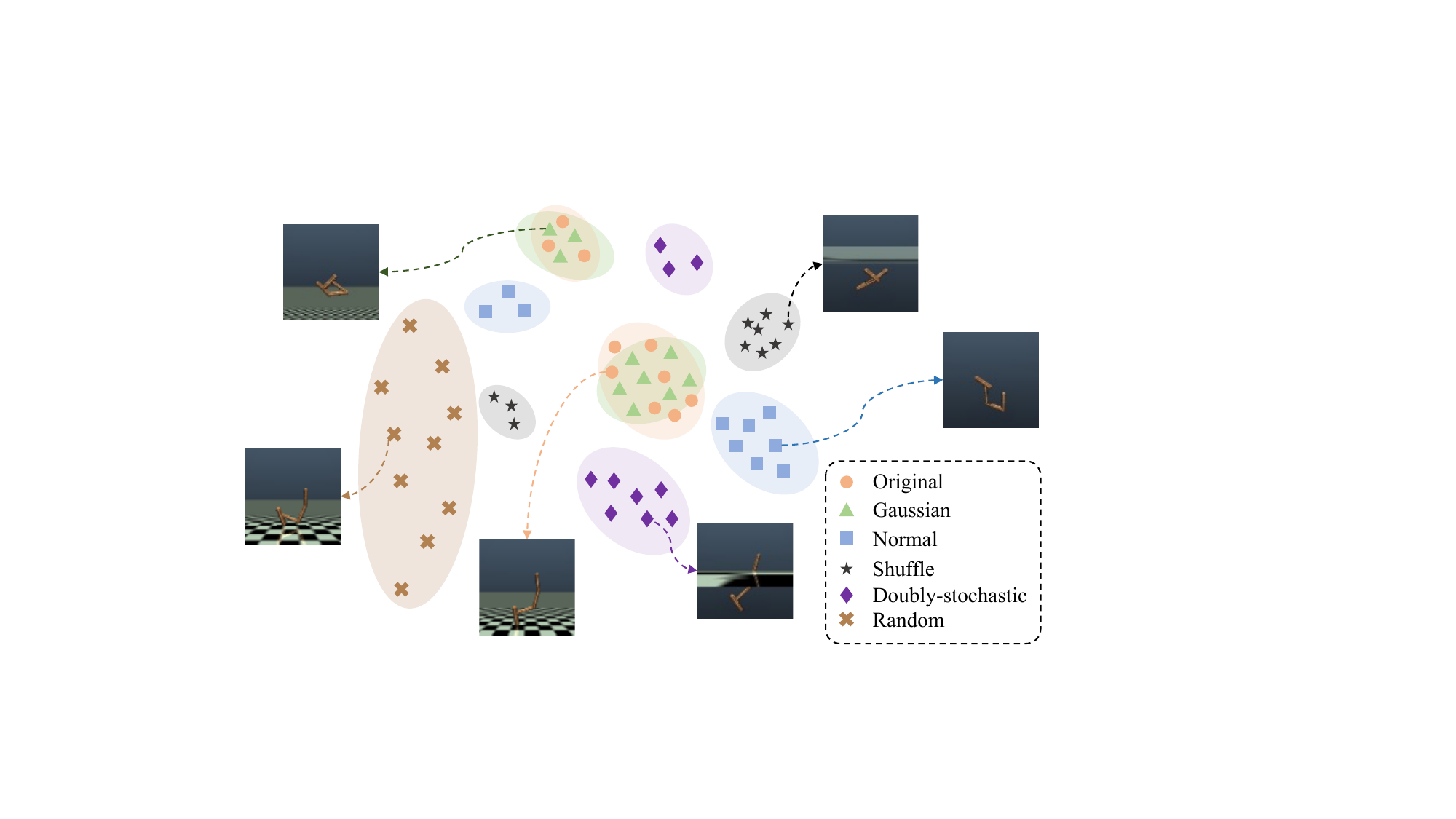}
\caption{The \textbf{simplified} t-SNE \cite{van2008visualizing} plot of states in collected trajectories. We selected 10 data points of each class from the original plot (\cref{f-full-tsne}) and displayed them using different colors and shapes. The arrows point to the rendered images using noisy underlying states, demonstrating the significant impact of noise on observations.}
\label{first}
\end{figure}

Imitation Learning (IL) \cite{schaal1999imitation,ross2010efficient,hussein2017imitation,osa2018algorithmic,zhang2023multi,gandhi2023eliciting} has made rapid progress in many challenging fields including game playing\cite{haliem2021learning,shi2023self}, financial trading \cite{peng2023valuation} and robotics \cite{duan2023structured,davis2023neuroceril,belkhale2023hydra}. In addition to the performance of a model, robustness and noise resilience are also important factors in IL algorithms designed for real-world problems. In this work, we focus on the problem of \textit{Learning from Noisy Demonstrations} (LND), whose name has been proposed by \cite{tangkaratt2020robust}. However, the specific problem setting in our work is quite different: the previous work learned from a mixture of expert and non-expert data, where the term ``imperfect" is more appropriate than ``noisy"; while we attempt to learn from expert data with various noise that are common in the real world.

The environment in our world is not a perfect simulator. Human experts may make mistakes; information encounters all kinds of noise during its transmission through a medium; sensors can suddenly malfunction. Even when we possess a perfect expert, the produced demonstrations may also become noisy when they reach the agent. Noise is inevitable in the real world. Therefore, making agents (also called imitators) robust to noise is crucial to the applications of IL methods.

Previous robust IL methods \cite{laskey2017dart,brown2020better,chen2021learning,wang2022adversarially} generally improve the robustness of the learned policy by injecting non-expert data or adversarially learned Gaussian noise into pure expert data. Some of which use additional ranking information. However, in most real-world scenarios, noisy expert data is more accessible than pure expert data, thus the LND setting that we propose is more realistic. There also exist methods \cite{tangkaratt2020robust,chen2023efficient,yuan2023good} learning from noisy expert data, however, they can handle either small scales of noise or only a single type of noise. 

To alleviate the above problems, we propose \textbf{D}enoised \textbf{I}mitation learning based on \textbf{D}omain \textbf{A}daptation (\textbf{DIDA}), which jointly learns a noise discriminator, a policy discriminator, and a feature encoder to learn representations that are task-related but domain-agnostic. Empirically, we find that \textit{DIDA can learn from fully noisy data without explicit knowledge of pure expert} and can solve a variety of noise, including additive noise and multiplicative noise.

Our main contributions are summarized as follows:

\textbf{1)} $\,\,$ We formally define a variety of noise for IL, including additive and multiplicative noise, and we theoretically analyze the limitations of GAIL in solving LND problems. \\
\textbf{2)} $\,\,$ We propose DIDA, a domain-adaptation-based Imitation Learning method, to imitate expert policy from noisy demonstrations in an adversarial manner, extracting domain-agnostic features and capturing the underlying expertise behind the noisy data. \\
\textbf{3)} $\,\,$ Experiments on 10 distinctive tasks with various types of noise on MuJoCo show the desirable performance of DIDA on LND problems.

\section{Related Work}
Many previous works focus on robust IL methods: DART \cite{laskey2017dart} injects noise into the expert's policy to reduce covariate shifts; RIL-Co \cite{tangkaratt2020robust} learns from a mixture of expert and non-expert demonstrations by co-pseudo-labeling; ARIL \cite{wang2022adversarially} designs an adversary to inject noise into real states, and jointly trains the agent's and adversary's policies; RIME \cite{chae2022robust} attempts to learn a robust policy by leveraging expert data from different MDPs. 
These methods require either expert policy or pure expert data; in contrast, \textit{DIDA only relies on fully noisy data}. D-REX \cite{brown2020better} and SSRR \cite{chen2021learning} use noisy policy to generate ranked trajectories and increase the robustness of the reward function respectively; other works \cite{burchfiel2016distance,brown2019extrapolating,wu2019imitation,zhang2021confidence} utilize ranking information among demonstrated trajectories by preference learning. Compared with the above methods, DIDA does not require additional ranking information, so it can work in situations where human ranking or querying is inaccessible.

DIDA adopts the framework of Adversarial Imitation Learning (AIL), where IL is interpreted as a divergence minimization problem \cite{ho2016generative,ghasemipour2020divergence,ke2021imitation}. Additionally, to obtain consistent representations, DIDA adopts the domain adversarial training paradigm, which has been involved in some IL methods \cite{stadie2017third,sharma2019third,liu2019state,cetin2021domain}. These methods treat imitation as a cross-domain problem and instruct the imitator to learn from expert demonstrations collected in a different domain. Our DIDA is closely related to these works, but has the following differences: 1) DIDA utilizes domain adaptation to learn from completely noisy data, while other methods learn from pure expert data. 2) We design a \textit{self-adaptive rate} and a \textit{domain adversarial sampling} (DAS) method to facilitate DIDA training in a self-adaptive way. 3) Many previous works require a \textit{random} dataset to help domain adaptation; we propose a new way to design such datasets that we call \textit{anchor buffer}, and illustrate its effect.

\section{Problem Setting}
\label{s-setting}

Suppose the expert buffer $\mathcal{B}_E=\{\tau_i^E\}_{i=1:T}$, where $\tau^E=\{(s_t,a_t)\}_{t=1}^H$. The noisy data can be generated by:
\begin{align}
    \tilde{s}_i=f_{\text{noise}}(s_i)&=f_i(A_is_i+B_i) \\
    \tilde{\tau}_i=f_{\text{noise}}(\tau_i)&=\{\tilde{s}_1,a_1,\dots,\tilde{s}_H,a_H\}
\end{align}
where $A\in\mathbb{R}^{n\times n}$, $s$ and $B$ are $n$-D vectors, $f(\cdot)$ is a transformation, which can be linear or nonlinear, and $f_{\text{noise}}$ is the noise operator. Subscript 
$i$ indicates that these components can be time-varying. We only consider the \textit{linear time-invariant (LTI)} \cite{gu2023mamba} noise on states: $\tilde{s}_i=f_{\text{noise}}(s_i)=As_i+B$. We ignore the noise on actions because the states usually have much higher dimensions than actions and thus are easier and more likely to be affected by noise. Our work primarily addresses two common noise in the signal transmission theory:

\textbf{Additive noise} can be added to the signal (expert data) and we formalize it as \textbf{\textit{Gaussian}} noise. We set $A$ to be the identity matrix, and the elements of $B$ are sampled from $\mathcal{N}(\mu,\sigma^2)$, where $\mu$ and $\sigma$ denote the expectation and standard deviation. 

\textbf{Multiplicative noise} is generally caused by imperfect channels and can be multiplied by the signal. Setting $B$ as zero vector, different choices of $A$ can lead to various noise: 

\textbf{1) \textit{Normal}:} Elements of $A$ are sampled from $\mathcal{N}(0,1)$. The resulting normal matrix multiplied by $s$ is equivalent to applying a linear transformation on the state feature space. 

\textbf{2) \textit{Doubly-stochastic}:} We set $A$ as a doubly-stochastic matrix, whose elements are all non-negative, and the sum of each row and column is equal to 1. Thus each element of $\tilde{s}$ is the weighted sum of all $n$ elements of $s$. The significance of studying \textit{Doubly-stochastic} noise is that if we can solve it, then its subclasses (such as permutation matrices and transition matrices) are also solvable.

\textbf{3) \textit{Shuffle}:} We set $A$ as a permutation matrix, which is a special case of the doubly-stochastic matrix. \textit{Shuffle} noise randomly permutes the dimensions of the state, which can simulate the disordered sensor data in reality.

In addition, we define the \textbf{\textit{combined}} noise, i.e., we uniformly and randomly divide the expert dataset into four parts, add the four noise defined above, and finally mix them up to form one dataset. 

We show the simplified t-SNE plot corresponding to different noisy data in \cref{first}, where the relative positions of different noise clusters in the figure reflect their properties. \textit{Original} and \textit{Random} are generated with the expert and random policies respectively and the other classes are obtained by adding different noise to \textit{Original}. \textit{Gaussian} surrounds \textit{Original}; \textit{Random} is the most dispersed; others have similar shapes to \textit{Original}, but differ in position.

\textbf{Definition of training data.} \,\, Let $\tau^E=\{(s^E_t,a^E_t)\}_{t=1}^H$ and $\tau^I=\{(s_t,a_t)\}_{t=1}^H$ denote the expert and imitator trajectories sampled with the expert policy $\pi_E$ and the imitator policy $\pi$ respectively. We define the expert buffer $\mathcal{B}_E=\{\tau^E_i\}_{i=1}^T = \{(s^E_t,a^E_t)\}_{t=1}^{T\times H}$ \footnote{For simplicity, we consider $\mathcal{B}_E$ as a set of $T\times H$ $(s,a)$ pairs.} and the imitator buffer $\mathcal{B}_I=\{\tau^I_i\}_{i=1}^T$ in the same way. Then we construct the noisy expert buffer $\tilde{\mathcal{B}}_E$:
\begin{align*}
    \tilde{\mathcal{B}}_E = f_{\text{noise}}(\mathcal{B}_E) &= \{f_{\text{noise}}(\tau^E_i)\}_{i=1}^T \\
    &= \{f_{\text{noise}}(s^E_t),a^E_t\}_{t=1}^{T\times H} \\
    &= \{(As^E_t+B,a^E_t)\}_{t=1}^{T\times H}
\end{align*} 
where noise operator $f_{\text{noise}}$ is either a single LTI noise or a mixture of multiple LTI noise. As shown in \cref{f-anchor}, by treating the expertise level and the noise level as the coordinate axes, we can place the datasets defined above into the four quadrants of the Cartesian coordinate system. For simplicity, we define the four quadrants of the coordinate axes as different domains $\mathcal{D}^{\text{EN}}$, $\mathcal{D}^{\text{RN}}$, $\mathcal{D}^{\text{RP}}$, and $\mathcal{D}^{\text{EP}}$, where (E, R, N, P) denote (\textbf{E}xpert, \textbf{R}andom, \textbf{N}oisy, \textbf{P}ure). Additionally, we define the expert domain $\mathcal{D}^{\text{E}}$ as $\mathcal{D}^{\text{EN}}\cup\mathcal{D}^{\text{EP}}$, and define the random domain $\mathcal{D}^{\text{R}}$, the noisy domain $\mathcal{D}^{\text{N}}$, and the pure domain $\mathcal{D}^{\text{P}}$ similarly.

\textbf{Learning from noisy demonstrations (LND).} \,\, Considering the Markov Decision Process (MDP) setting, we assume that the expert $\pi_E$ and the imitator $\pi$ share the same dynamics $\mathcal{M}=(\mathcal{S},\mathcal{A},\mathcal{P},r,\gamma,p_0)$, where $\mathcal{S}$ is the state space, $\mathcal{A}$ is the action space, $\mathcal{P}\colon\mathcal{S}\times\mathcal{A}\rightarrow\mathcal{S}$ is the transition function, $r\colon\mathcal{S}\times\mathcal{A}\times\mathcal{S}\rightarrow\mathbb{R}$ is the reward function, $\gamma\in[0,1)$ is the discount factor, and $p_0$ is the initial state distribution. According to Proposition 3.1 in \cite{ho2016generative}, the occupancy measure $\tilde{\rho}_e$ derived from the noisy expert buffer $\tilde{\mathcal{B}}_E$ refers to a unique corresponding noisy expert policy $\tilde{\pi}_E$, and the same is true for the anchor domain $(\tilde{\mathcal{B}}_A,\tilde{\rho}_A,\tilde{\pi}_A)$ that will be discussed later in \cref{anchor buffer}. The corresponding dynamics of $\tilde{\mathcal{B}}_E$ is $\mathcal{M}_n=(\mathcal{S}_n,\mathcal{A}_n,\mathcal{P}_n,r_n,\gamma,p_0^n)$, where $\mathcal{A}_n=\mathcal{A}$ as we ignore the noise on actions. For all $s,s'\in\mathcal{S}$ and $\tilde{s}=f_{\text{noise}}(s),\tilde{s}'=f_{\text{noise}}(s')\in\mathcal{S}_n$, we have $p_0(s)=p_0^n(\tilde{s})$, $\mathcal{P}(s,a,s')=\mathcal{P}(\tilde{s},a,\tilde{s}')$ and $r(s,a)=r_n(\tilde{s},a)$. The main challenge in LND is that the states $s\in\mathcal{S}_n$ and actions $a\in\mathcal{A}_n$ in the noisy expert's MDP $\mathcal{M}_n$ do not necessarily match the states $s\in\mathcal{S}$ and actions $a\in\mathcal{A}$ in the imitator's MDP $\mathcal{M}$. 

\section{Domain Adaptation for Noisy Expert Data}
Although AIL has shown promising performance on various imitation learning tasks, it remains unclear whether it can solve the problem of LND where the expert and the imitator have different domains caused by noise. In the first part, we theoretically analyze the limitations of AIL methods (especially GAIL) in solving the LND problems. This explains why we need to introduce domain adaptation techniques. In the second part, we introduce the anchor buffer, a technique commonly used in domain adaptation methods, and analyze it in the context of LND.

\subsection{Why do we need domain adaptation?}
AIL can hardly solve LND problems, for it essentially attempts to justify expert data rather than exploring the logic behind expert behavior \cite{brown2019extrapolating}. To prove the limitations of the AIL methods in dealing with the LND problem, we start with GAIL and propose the following proposition:
\begin{proposition}
\label{pro1}
    (Limitation of GAIL) Given expert data with LTI noise and no perfect information, GAIL can only solve small-scale Gaussian noise and cannot handle other LTI noise, proof in \cref{proof}.
\end{proposition}
To address the Imitation Learning problem with different domains between the expert and the imitator under the framework of AIL, we need to utilize consistent representations, which can be achieved by domain adaptation techniques. Domain adaptation methods aim to map the noisy expert domain and the noiseless imitator domain to the same latent space, thus reducing the LND problem to a simple IRL case.

\subsection{Anchor buffer}\label{anchor buffer}
Anchor buffer is a commonly used design in domain adaptation methods, which is also used in DIDA. We provide insights into understanding the necessity of the anchor buffer and analyze how the anchor buffer works under the LND setting. 

\begin{figure}[h]
    \centering
    \includegraphics[width=0.6\columnwidth]{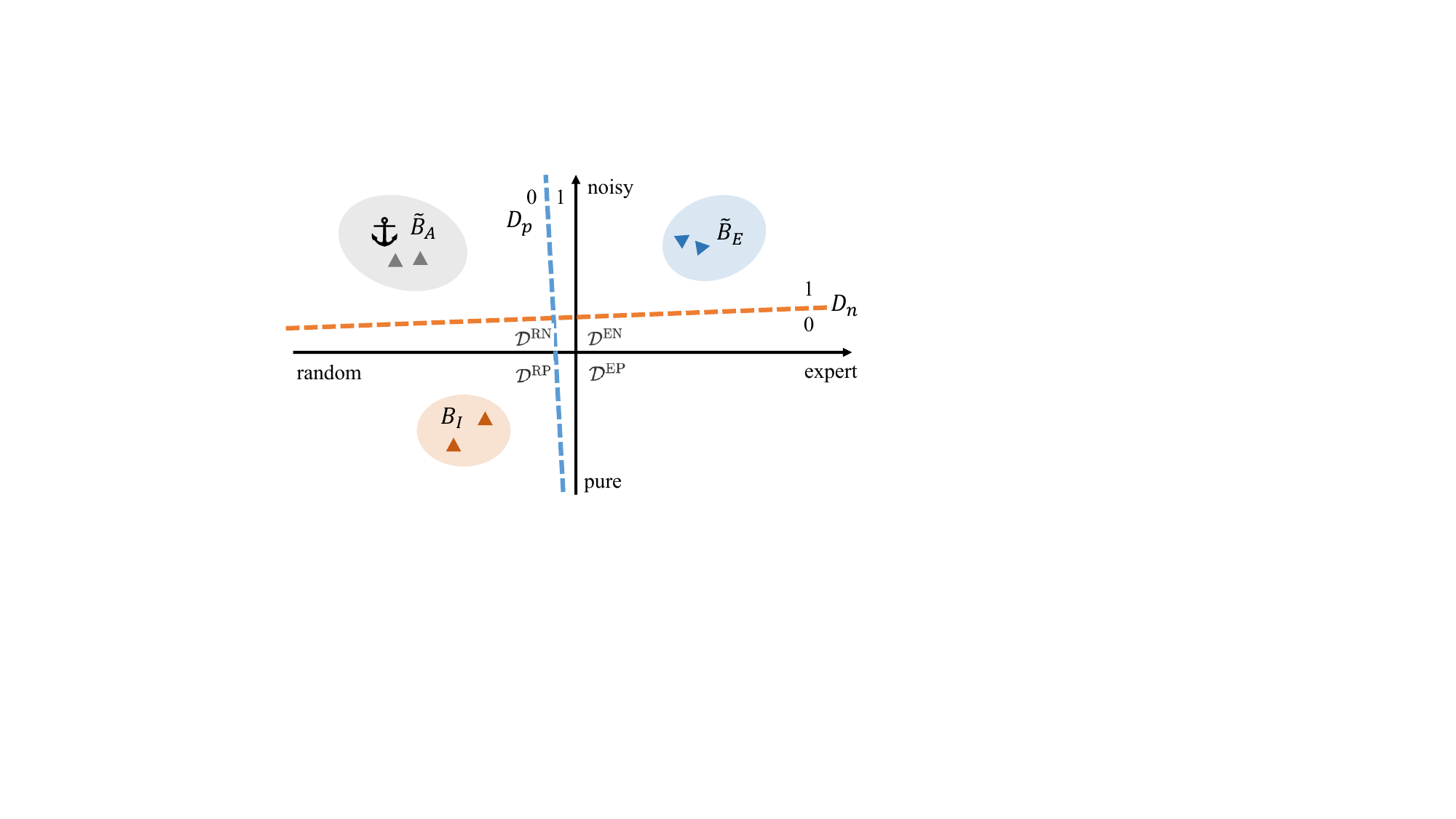}
    \caption{We consider the expertise level and the noise level of data as the x-axis and the y-axis, respectively. $\tilde{\mathcal{B}}_E$ and $\mathcal{B}_A$ are located at the diagonal position.}
    \label{f-anchor}
\end{figure}

Considering the simplest case in \cref{f-anchor}, where the discriminator's classification boundary can be seen as a straight line, $\tilde{\mathcal{B}}_E$ and $\mathcal{B}_I$ can be separated easily by any classification boundary with slope $k\in(-\infty,0]$. So we need to construct a set of data within the noisy expert domain to aid in training, and we call such a dataset an \textbf{anchor buffer}. The anchor buffer acts as the bridge to connect the other two buffers: $\tilde{\mathcal{B}}_A$ and $\tilde{\mathcal{B}}_E$ belong to $\mathcal{D}^{\text{N}}$ with the same noise level, and training a policy discriminator $D_p$ with these two buffers allows $D_p$ to categorize the data based on the expertise level only; $\tilde{\mathcal{B}}_A$ and $\tilde{\mathcal{B}}_I$ belong to $\mathcal{D}^{\text{R}}$ with the same expertise level, and thus guarantees a noise discriminator $D_n$ to rely only on the noise level of the data for classification.

In addition, there are certain problems with the way that several previous domain adaptation methods construct their anchor buffers. For instance, DisentanGAIL \cite{cetin2021domain} collects prior data by executing random behavior in both expert and imitator domains. TPIL \cite{stadie2017third} also designs a similar dataset by executing random policies in the expert domain. The above two methods need interactions within the expert's domain to collect prior data and we refer to this generation method of anchor buffer as \textit{anchor-buffer-random}. In many IL scenarios, however, the imitator only has access to the expert demonstrations and cannot interact within the expert domain. Therefore, it is unrealistic to collect data through extra interactions. To address this problem, DIDA proposes a new method for constructing anchor buffers without pre-collected random data from the expert domain, which will be introduced in the next section.

\begin{figure*}[t]
    \centering
    \includegraphics[width=0.85\textwidth]{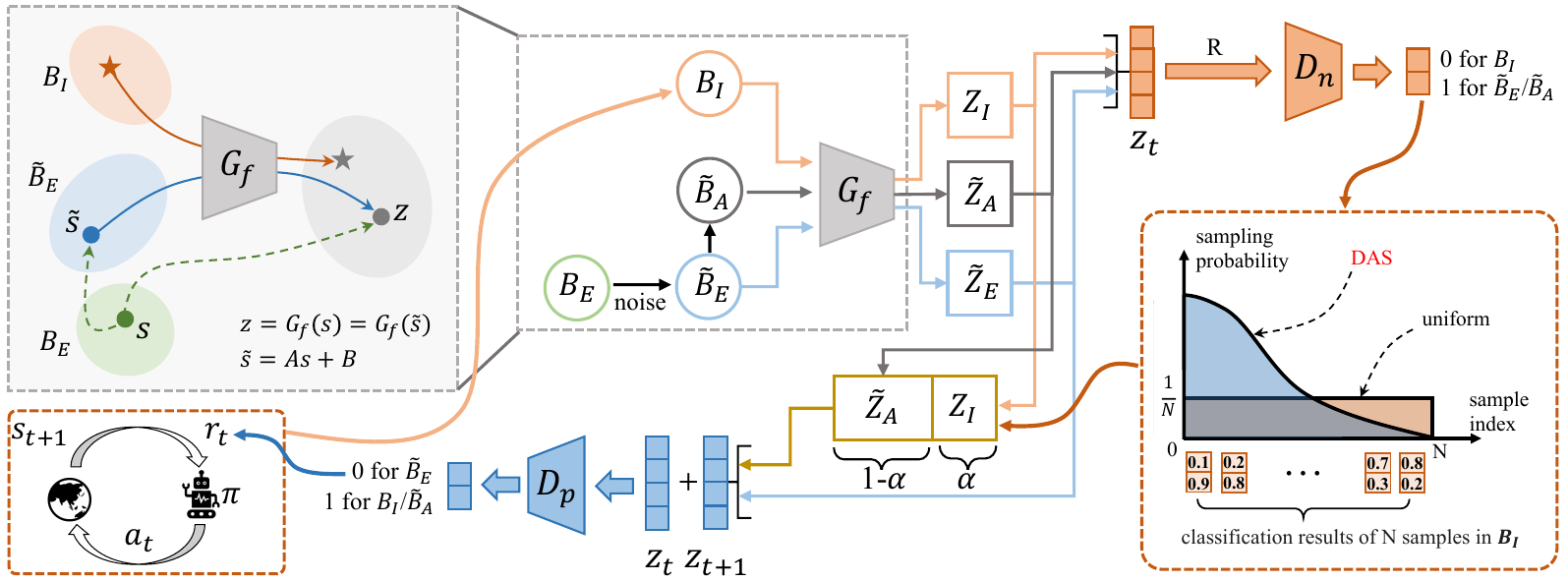}
      \caption{The main framework of the DIDA method. We design the feature encoder $G_f$ to map state $s$ into embedding $z$. We sample batches of size $N$ in the imitator buffer $\mathcal{B}_I$, the noisy anchor buffer $\tilde{\mathcal{B}}_A$, and the noisy expert buffer $\tilde{\mathcal{B}}_E$ (defined in \cref{s-setting}) and feeding them into $G_f$ to get embeddings $Z_I$, $\tilde{Z}_A$, and $\tilde{Z}_E$. Noise discriminator $D_n$ judges the noise level of all embeddings and generates binary classification results. $R$ denotes the gradient reversal layer. We apply the technique of domain adversarial sampling (DAS, defined in \cref{s-DAS}) to compute the confusion probability distribution $P_{das}(z^I)$ over $Z_I$ based on the classification error probability of the embeddings. We take $\alpha N$ embeddings from $Z_I$ based on $P_{das}$ and replace the $\alpha N$ embeddings in $\tilde{Z}_A$ with them randomly, where $\alpha$ is the adaptive rate from \cref{s-adaptive-rate}. Policy discriminator $D_p$ judges the expertise level of $Z_{\text{mix}}$ and $\tilde{Z}_E$.} 
      \label{f-main}
\end{figure*}

\section{A Practical Implementation of DIDA}
In this section, we will detail our design of a denoised imitation learning method, DIDA. In \cref{s-components}, we introduce the components and the optimization objective of DIDA. In \cref{s-DAS}, we propose Domain Adversarial Sampling (DAS), which is an imbalanced sampling technique. In \cref{s-adaptive-rate}, we explain the design idea of a self-adaptive rate.

\subsection{Components and Optimization objective}
\label{s-components}
Inspired by the domain adaptation theory \cite{ben2006analysis}, we aim to learn a representation that is not discriminative between the noisy domain $\mathcal{D}^{\text{N}}$ and the pure domain $\mathcal{D}^{\text{P}}$, but can still be used to perform IL tasks well. To achieve this goal, we design a feature encoder $G_f$ to extract the state embedding $z_t=G_f(s_t)$, a noise discriminator $D_n$, and a policy discriminator $D_p$. In each iteration, we sample from $\mathcal{B}_I$, $\tilde{\mathcal{B}}_A$, and $\tilde{\mathcal{B}}_E$ to get corresponding state batches $S_I=\{s^I_i\}_{i=1}^N$, $\tilde{S}_A=\{s^A_i\}_{i=1}^N$, and $\tilde{S}_E=\{s^E_i\}_{i=1}^N$, respectively. Then we use $G_f$ to map these batches into the imitator embeddings $Z_I=\{z^I_i\}_{i=1}^N$, the noisy anchor embeddings $\tilde{Z}_A=\{z^A_i\}_{i=1}^N$ and the noisy expert embeddings $\tilde{Z}_E=\{z^E_i\}_{i=1}^N$, where $z^*_i=G_f(s^*_i)$. DIDA proposes two adversarial training branches: \textbf{Policy branch} takes policy $\pi$ as the generator and combines $G_f$ with $D_p$ to distinguish whether $z_t$ belongs to the anchor domain $\mathcal{D}^{\text{R}}$; \textbf{Noise branch} takes $G_f$ as the generator, and uses $D_n$ to distinguish whether $z_t$ belongs to the noisy domain $\mathcal{D}^{\text{N}}$. Well-trained discriminators should be orthogonal to the coordinate axes in \cref{f-anchor}, and we expect $D_p$ and $D_n$ to make decisions independently. 

\textbf{Optimization objective of DIDA.} \,\, For simplicity, we denote $\mathbb{E}_{s\sim\pi}[f(G_f(s))]=\mathbb{E}_{\pi}[f(z)]=\mathbb{E}_{\mathcal{B}_I}[f(z)]$ where $f$ can be any function, and abbreviate $(z_t,z_{t+1})$ as $\sigma_t$. Aiming to derive the optimization objective for the entire DIDA framework, we start with the policy branch and then introduce the noise branch. The optimization objective of the policy branch is:
\begin{align}
    \max_{\pi}\min_{D_p}\mathcal{L}_p &= \mathbb{E}_{\pi\cup\tilde{\pi}_r}[\log(1-D_p(\sigma_t))] \\
    &\qquad + \mathbb{E}_{\tilde{\pi}_e}[\log D_p(\sigma_t)] \\
    \label{e6}
    &= \mathbb{E}_{\pi\cup\tilde{\pi}_r\cup\tilde{\pi}_e}[L_p(D_p(\sigma_t),p_{l_t})] \\
    \label{e7}
    &= \mathbb{E}[L_p(D_p(\sigma_t),p_{l_t})]
\end{align}
where $L_p$ is the policy loss function, $l_{t}^p=\mathbb{I}(s_t\in\tilde{\mathcal{B}}_A\cup\mathcal{B}_I)$ is the policy domain label to separate $\mathcal{D}^{\text{R}}$ and $\mathcal{D}^{\text{E}}$. $\mathbb{I}$ is the indicator function. \cref{e6} is computed over all datasets, so we drop its subscript for simplicity. To obtain domain-agnostic embeddings, we introduce the following mutual information term:
\begin{align}
    \max_{\pi}\min_{D_p}\mathcal{L}_p = &\mathbb{E}[L_p(D_p(\sigma_t),l_{t}^p)] \\
    \label{e-MI}
    &\text{s.t.} \,\, MI(z,n_l)=0
\end{align}
where the noise domain label $l_{t}^n=\mathbb{I}(s\in\tilde{\mathcal{B}}_E\cup\tilde{\mathcal{B}}_A)$ is the indicator to separate $\mathcal{D}^{\text{N}}$ and $\mathcal{D}^{\text{P}}$. The mutual information (MI) term forces $G_f$ to learn embedding that is not discriminative between the noisy domain $\mathcal{D}^{\text{N}}$ and the pure domain $\mathcal{D}^{\text{P}}$. DIDA instantiates the MI term with the noise discriminator $D_n$ to maximize domain confusion:
\begin{align}
    \max_{\pi}\min_{D_p}\max_{D_n}\mathcal{L}_p+\mathcal{L}_n = \mathbb{E}[&L_p(D_p(\sigma_t),l_{t}^p) \\
    + &L_n(D_n(z_t),l_{t}^n)]
\end{align}
where $L_n$ is the noise loss function and can have a different form than $L_p$. We apply the gradient flipped technique from \cite{ganin2016domain} to tackle the conflict of update goals between the noise discriminator and the feature encoder. Then the optimization objective becomes:
\begin{align}
    \max_{\pi}\min_{D_p,D_n,G_f}\mathcal{L}_p+\mathcal{L}_n = \mathbb{E}[L_p(D_p(\sigma_t),l_{t}^p) \\
    + \lambda L_n(D_n(R(G_f(s_t))),l_{t}^n)]
\end{align}
The gradient reversal layer $R$ acts as an identity function in the forward propagation, while flips the sign of gradients in the backward propagation. Domain weight $\lambda$ controls the trade-off between the two loss terms, which can be computed as \cref{alg1}. According to \cite{cetin2021domain}, the information contained in $z$ can be divided into the \textit{domain information} and the \textit{goal-completion information}. With the model converging, the correlation between the embedding $z$ and the noise label $l_t^n$ is minimized. In other words, domain information is eliminated as much as possible, leaving only the goal-completion information which is useful to the policy branch training. 

\textbf{New construction method of anchor buffer.}\,\, We randomly shuffle the state-action pairs of the noisy expert buffer $\tilde{\mathcal{B}}_E$ to obtain the noisy random buffer (mentioned as the anchor buffer) $\tilde{\mathcal{B}}_A=f_{\text{shuffle}}(\tilde{\mathcal{B}}_E)$. We refer to this generation method as \textit{anchor-buffer-shuffle}. We conduct ablation experiments in \cref{exp-ablation} to explore and compare the effects of different anchor buffer construction methods on experimental performance.

\begin{figure}[t]
    \centering
    \includegraphics[width=\columnwidth]{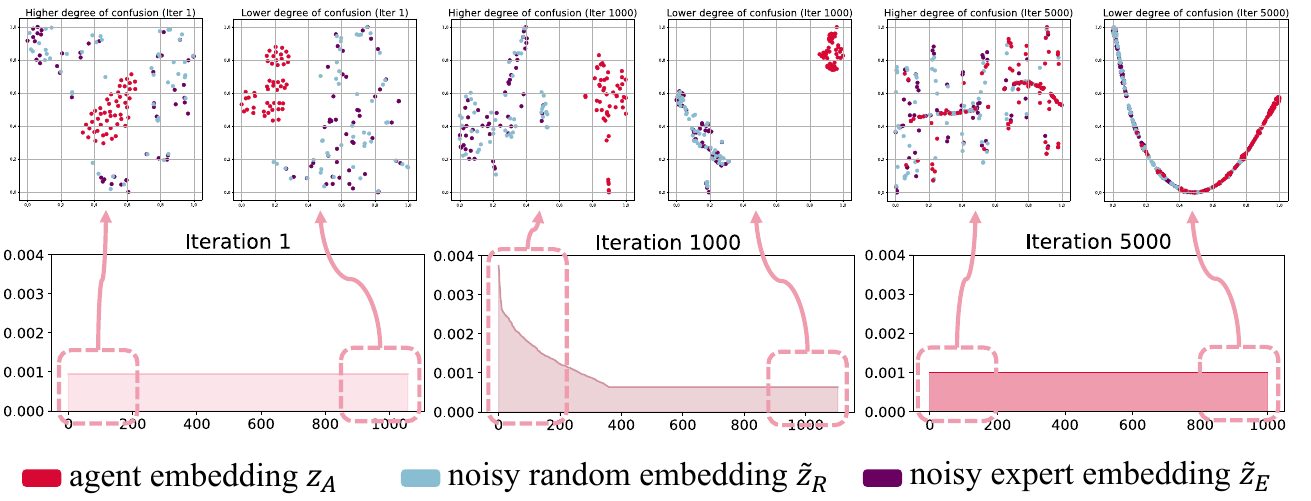}
    \caption{\textbf{Top}: The confusion probability $P_{das}$ of $Z_A$ at iter-1, iter-1000, and iter-5000 are shown from left to right, sorting the samples with their degrees of confusion decreasing from left to right.
    \textbf{Bottom}: The t-SNE plots of embeddings from $Z_A$, $\tilde{Z}_R$, and $\tilde{Z}_E$ at iter-1000. From left to right corresponds to embeddings with the highest, middle, and lowest degrees of confusion at iter-1000, respectively. More detailed t-SNE plots in \cref{appendix-confusion-tsne}.}
    \label{f-distribution-change}
\end{figure}

\subsection{Domain Adversarial Sampling (DAS)}
\label{s-DAS}

The imitator embeddings $Z_I$ is an \textit{imbalanced} dataset for the distribution of domain information is different across samples, which can be justified by the lower center figure of \cref{f-distribution-change}. $D_n$ distinguishes whether an embedding is from $\mathcal{D}^{\text{N}}$ or $\mathcal{D}^{\text{P}}$ based on the domain information, so the classification accuracy can be used to measure the amount of domain information in the embedding. High-accuracy samples are detrimental to policy branch training because they contain plenty of domain information. Therefore, we need to design an imbalanced sampling method rather than uniform sampling to extract samples with less domain information in $Z_I$ and use these samples to train the policy discriminator. We define the following confusion probability:
\begin{align}
    P_{das}(z^I_i) &= \frac{p_i}{\sum_{j=1}^N p_j} \\
    \text{where}\quad p_i&=P(D_n(z^I_i)=1),\forall z^I_i\in Z_I
\end{align}
We refer to $P_{das}(z^I_i)$ as the \textit{degree of confusion} of $z^I_i$. 
We take $\alpha N$ embeddings from $Z_I$ based on $P_{das}$ and replace the $\alpha N$ embeddings in $\tilde{Z}_A$ with them randomly. Then we use the obtained mixed embeddings $Z_{\text{mix}}$ to train the policy discriminator.

As shown in the bottom rows of \cref{f-distribution-change}, the confusion probabilities of iter-1 and iter-5000 are both nearly uniform distributions. Because no prior knowledge is introduced at iter-1, $D_n$ classifies all embeddings into the same class with approximately equal probabilities and the domain confusion is maximized. At iter-5000, all domain information is eliminated from embeddings, at which point $D_n$ is confused about the domains and performs similar to in iter-1. In the top rows of \cref{f-distribution-change}, samples with greater degrees of confusion are closer to each other in different clusters.

\subsection{Self-Adaptive Rate (SAR)}
\label{s-adaptive-rate}

The $Z_{\text{mix}}$ used for training the policy discriminator is composed of $Z_I$ and $\tilde{Z}_A$. The proportion of embeddings from $Z_I$ needed at different stages of training is not the same, so we design the adaptive rate $\alpha$ to control this proportion. We define $p_{\text{acc}}$ as the classification accuracy of $D_n$ on all data, and define $\alpha$ as a function of $p_{\text{acc}}$:
\begin{align}
    \alpha &= f_{\alpha}(p_{\text{acc}}) \\
    \label{e-adaptive}
    &= \mathbb{I}(p_{\text{acc}}>p)\frac{1-p_{\text{acc}}}{1-p} + \mathbb{I}(p_{\text{acc}}\leq p)\frac{p_{\text{acc}}}{p}
\end{align}
where $f_{\alpha}$ is the adaptive function. $p$ is a hyperparameter and DIDA sets it as $\frac{2}{3}$ (can also be denoted as SAR-$\frac{2}{3}$). The value comes from the proportion of noisy data in all data $(|\tilde{Z}_E|+|\tilde{Z}_A|) / (|\tilde{Z}_E|+|\tilde{Z}_A|+|Z_I|)=\frac{2}{3}$. We perform ablation in \cref{exp-ablation} for different settings of $p$. The adaptive rate behaves as a linear function in both $[0,p]$ and $[p,1]$. $\alpha\in[0,1]$ reaches its maximum value of $1$ at $p_{\text{acc}}=p$, and reaches the minimum value of $0$ at $p_{\text{acc}}=0$ or $1$. 

Our design concerning $f_{\alpha}$ is explained below. Flipping the output of a classifier with $p_{\text{acc}}=1$ turns it into a classifier with $p_{\text{acc}}=0$, demonstrating that achieving low accuracy still requires much domain information. Thus too high or too low classification accuracy of $D_n$ indicates that $G_f$ fails to remove the domain information from embeddings. In this case, we reduce the proportion of $Z_I$, and use more of $\tilde{Z}_A$ and $\tilde{Z}_E$ as input which can prevent $D_p$ from gaining the ability to classify through domain information.

\begin{figure}[t]
    \centering
    \includegraphics[width=\columnwidth]{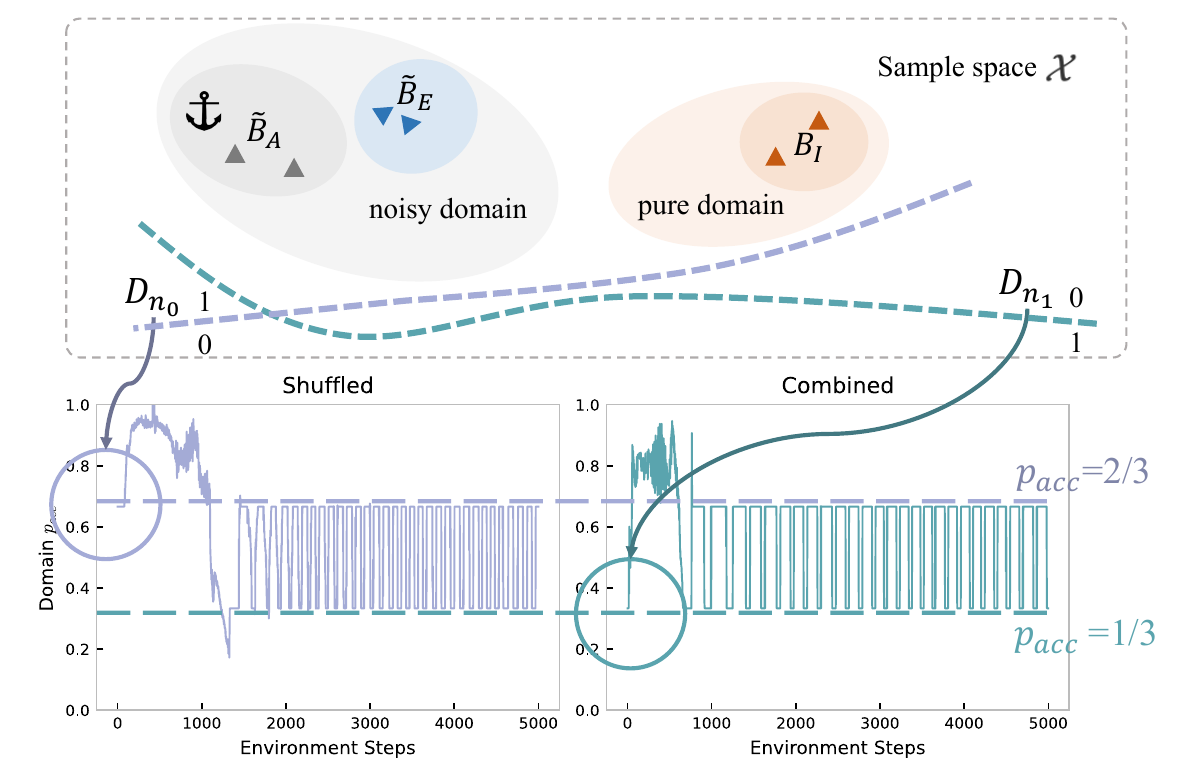}
    \caption{Curve of change in $p_{\text{acc}}$. \textbf{Top}: $\mathcal{B}_I$, $\tilde{\mathcal{B}}_A$ and $\tilde{\mathcal{B}}_E$ only occupy a tiny portion of the $n$-dimensional state space $\mathcal{X}$. When $D_n$ is randomly initialized, the classification hyperplane has a high probability of classifying all datasets into the same category, resulting in an accuracy of $\frac{2}{3}$ (e.g. $D_{n_0}$) or $\frac{1}{3}$ (e.g. $D_{n_1}$).\textbf{Bottom Left}: The initial value is $p_2$, and it fluctuates between $p_1$ and $p_2$ in the late stage of training. \textbf{Bottom Right}: The initial value is $p_1$ and it also fluctuates between $p_1$ and $p_2$.}
    \label{f-fluctuate}
\end{figure}

\cref{f-fluctuate} shows two intriguing phenomena: \textbf{(1)} The value of $p_{\text{acc}}$ is mostly $p_1=\frac{1}{3}$ and $p_2=\frac{2}{3}$ at iter-1. \textbf{(2)} $p_{\text{acc}}$ fluctuates rapidly and regularly between $p_1$ and $p_2$ in the late stage of training. The top plotting in \cref{f-fluctuate} explains the first phenomenon: after randomly initializing the parameters, $D_n$ will most likely classify the three datasets into the same category $0$ or $1$ ($0$ corresponds to $p_{\text{acc}}=p_1$, and $1$ corresponds to $p_{\text{acc}}=p_2$). Thus we believe that $p_1$ and $p_2$ are special accuracy values that correspond to the random classification results without any prior knowledge. Then we attempt to explain the second phenomenon. With the model converging, we expect $G_f$ to eliminate all domain information of the state embeddings, and thus satisfy $MI(z,l_t^n)=0$. At this time, $D_n$ classifies randomly similarly to the iter-1 moment, thus $p_{\text{acc}}$ fluctuates between $p_1$ and $p_2$. Results in \cref{f-distribution-change} also verify these phenomena.

\begin{table*}[t]
\renewcommand{\arraystretch}{1.25}
\vskip 0.15in
\caption{Test returns of DIDA and the other baseline methods on two MuJoCo environments with five noise. Each combination of environment and noise type is considered as a task (12 tasks in total). The table reports the mean (higher is better) and standard deviation of the return evaluated on all tasks for each method over three trials.}
\begin{center}
\begin{scriptsize}
\begin{tabular}{c|l|ccccccc}
\toprule
\textbf{Env}  &  \textbf{Noise}  &   \textbf{DIDA}(ours)  &  \textbf{GWIL}  &  \textbf{SAIL}  &  \textbf{TPIL}  &  \textbf{GAIL}  &  \textbf{BC} \\
\midrule
  &  None  &    1311.0$\pm$518.6  &  694.4$\pm$300.6  &  -18.4$\pm$1.1  &  1506.3$\pm$795.2  &  \textbf{1708.4$\pm$436.8}  &  1605.8$\pm$489.0  \\
  &  Shuffle  &    \textbf{2248.1$\pm$475.8}  &     822.6$\pm$224.7  &  21.9$\pm$1.0  &  573.9$\pm$354.2  &  316.1$\pm$463.0  &  76.1$\pm$98.6  \\
Hopper  &  Doubly-stochastic  &    \textbf{1598.6$\pm$939.3}  & 955.4$\pm$256.1 &  0.7$\pm$0.1  &  974.1$\pm$913.7  &  447.4$\pm$165.7  &  1.7$\pm$3.0  \\
(Expert:1813.6$\pm$590.5)  &  Normal  &    \textbf{1271.8$\pm$618.2}  &    675.7$\pm$367.4   &  -95.3$\pm$15.4  &  560.9$\pm$344.6  &  458.6$\pm$348.6  &  4.4$\pm$1.1  \\
  &  Gaussian  &    \textbf{2284.4$\pm$807.2}  &  873.2$\pm$248.5   &  1021.5$\pm$0.3  &  573.1$\pm$337.4  &  454.9$\pm$303.6  &  89.6$\pm$31.3  \\
  &  Combined  &    \textbf{2340.6$\pm$808.3}  &  939.6$\pm$187.0&  -91.8$\pm$44.2  &  483.7$\pm$237.9  &  402.2$\pm$156.7  &  141.9$\pm$69.4  \\
\midrule
  &  None  &   120.8$\pm$1.8  &   5.0$\pm$11.9 &  -71.4$\pm$55.9  &  117.5$\pm$6.1  &  119.6$\pm$3.7  &  \textbf{122.8$\pm$2.4}  \\
  &  Shuffle  &    \textbf{93.0$\pm$36.6}  &   1.5$\pm$18.1 &  -6.6$\pm$7.9  &  31.9$\pm$41.4  &  38.0$\pm$14.5  &  17.0$\pm$6.2  \\
Swimmer  &  Doubly-stochastic  &    \textbf{74.4$\pm$18.0}  &  -1.6$\pm$13.6 
 &  1.7$\pm$18.1  &  -1.1$\pm$14.7  &  69.8$\pm$8.3  &  -46.6$\pm$45.0  \\
(Expert:122.8$\pm$1.7)  &  Normal  &    \textbf{39.8$\pm$13.5}  &   0.8$\pm$12.2  &  -16.9$\pm$6.4  &  27.9$\pm$15.4  &  21.4$\pm$23.5  &  17.9$\pm$17.3  \\
  &  Gaussian  &    \textbf{72.0$\pm$35.5}  &  -3.1$\pm$11.0&  -6.7$\pm$8.8  &  57.5$\pm$23.7  &  43.6$\pm$38.3  &  36.1$\pm$5.9  \\
  &  Combined  &    \textbf{62.2$\pm$12.7}  &  -3.6$\pm$15.1&  17.5$\pm$12.7  &  37.6$\pm$31.7  &  49.3$\pm$12.1  &  32.0$\pm$4.6 \\
\bottomrule
\end{tabular}
\end{scriptsize}
\end{center}
\vskip -0.1in
\label{table1}
\end{table*}

\section{Experiments}
We conduct several experiments to answer the following scientific questions: 

\textbf{(1) Can} DIDA achieve excellent performance in different environments and under various noise? 

\textbf{(2) How} do the different components of DIDA contribute to the final performance? 

\textbf{(3) What} are the effects of different types of noise on imitation learning? Can we extract useful information about imitating noisy demonstrations?

\textbf{Datasets Preparation.}\,\, We test the ability of DIDA and other baseline methods with two control tasks from MuJoCo \cite{todorov2012mujoco}: \textit{Hopper} and \textit{Swimmer}. Firstly, we run \textbf{PPO} \cite{schulman2017proximal} on each environment for 1000 episodes to obtain the expert policy $\pi^E$, compute its test return for 100 episodes, and record its performance. Then we collect 50 episodes of expert buffer $\mathcal{B}_E$ with the expert policy. Finally, to simulate the noise introduced during data transmission, we add various noise on $\mathcal{B}_E$ to create corresponding noisy expert buffers. It is worth mentioning that we set $\mu=0$ and $\sigma=0.1$ for \textit{Gaussian} noise; we split $\mathcal{B}_E$ into four parts randomly and evenly, add four different noise to each part, and combine them to create the buffer of \textit{Combined} noise.

\textbf{Baselines.}\,\, We compare our method to 5 baselines detailed in \cref{baselines}. \textbf{BC} takes a supervised learning way to learn a policy that maps states to actions using expert demonstrations. IRL methods like \textbf{GAIL} \cite{ho2016generative} adopt the idea of GAN \cite{goodfellow2014generative} and learn a policy distribution by generating state-action pairs that are indistinguishable from those of an expert policy, \textbf{TPIL} \cite{stadie2017third} addresses the domain gap by aligning the domains of the expert and the learner in the third-person imitation learning setting, while \textbf{GWIL} \cite{fickinger2021cross} uses the Gromov-Wasserstein distance to align and compare states between the different spaces of the agents. Moreover, \textbf{SAIL} \cite{liu2019state} is a state adaptation-based algorithm that addresses the problem of different dynamics models between the imitator and the expert. TPIL, GWIL and SAIL all utilize domain adaptation just as DIDA. Since in our setting noise is only added to states, it's worth exploring if SAIL can work well on LND problems.

\subsection{Can DIDA achieve excellent performance in different environments and under various noise?}
\label{s-baseline}
We evaluate DIDA and other baseline methods on tasks across different environments and noise. Results in \cref{table1} shows that DIDA significantly outperforms other baselines under all noise conditions, and achieves comparable results to the best baseline on the pure expert cases, demonstrating the effectiveness of domain adaptation between agent domain and noise domain. DIDA generally performs better on Hopper than on Swimmer, especially achieving better than expert performance under \textit{Shuffle}, \textit{Guassian} and \textit{Combine} noise. BC simply learns a mapping from the noisy state space $\mathcal{S}_n$ to the action space $\mathcal{A}_n$ using supervised learning paradigm. The learned policy has almost no noise resistance, so although it performs well in the pure expert case, it performs much worse in all noisy cases. As a domain adaptation method, TPIL has a certain degree of noise resistance. However, due to the lack of self-adaptive designing, its performance is inferior to DIDA. GAIL's performance is roughly comparable to TPIL's, as it simply attempts to align expert data rather than extracting expertise knowledge behind expert behavior. GWIL's performance is quite good on Hopper, but it gets relatively low returns on Swimmer. SAIL also cannot handle noisy demonstrations very well. These results show that our proposed method, DIDA, can achieve robust good performance on almost all designed tasks compared to other domain adaptation methods.

\begin{figure}[t]
    \centering
    \includegraphics[width=0.5\textwidth]{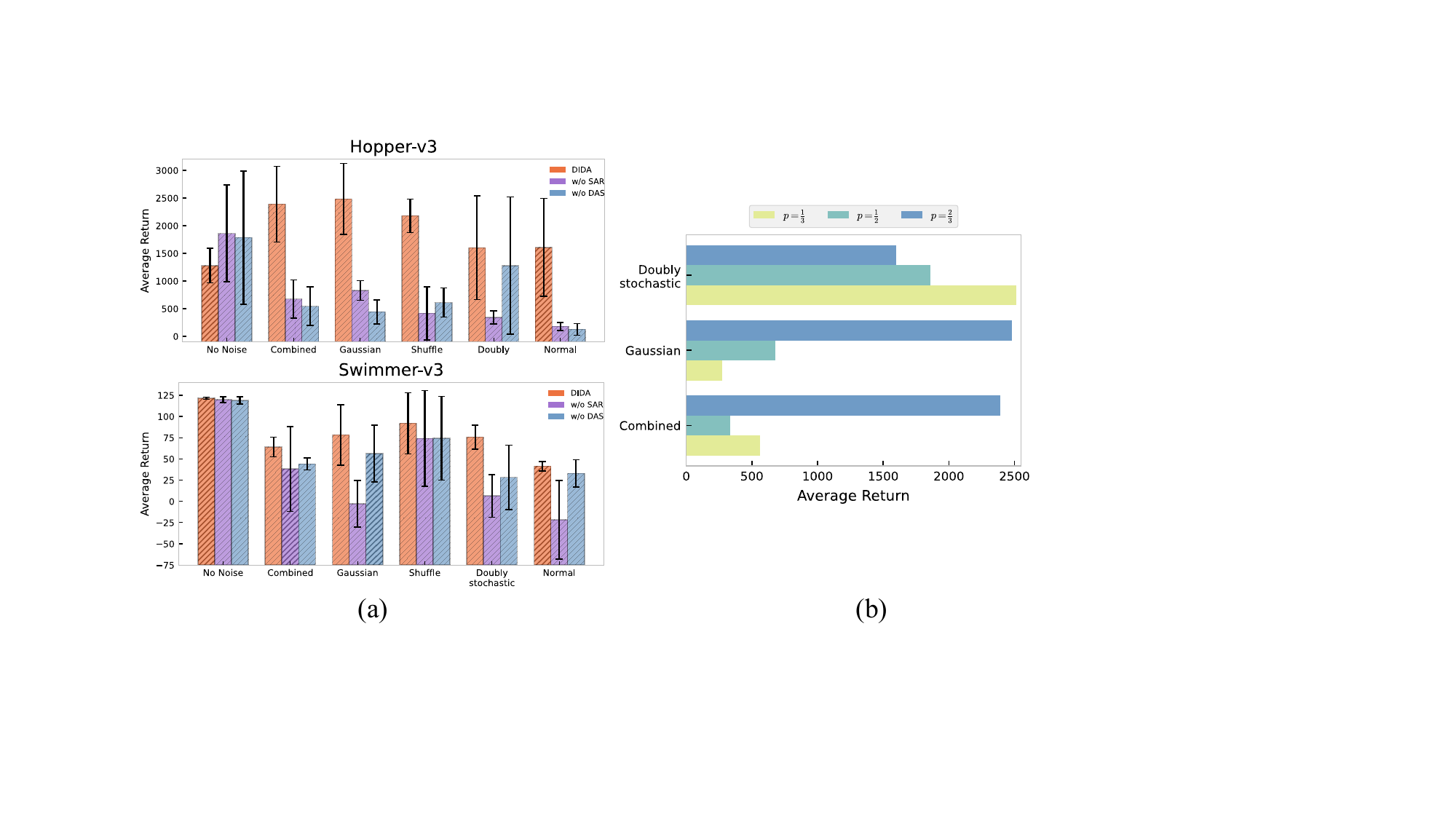}
    \caption{\textbf{(a)}: Results of ablation studies of SAR and DAS with all noises on Hopper and Swimmer. \textbf{(b)}: Performances of ablation studies on different settings of $p$ in SAR.} 
    \label{ablation-single}
\end{figure}

\subsection{How do the different components of DIDA contribute to final performance?}
\label{exp-ablation}

To evaluate the contribution of different components of DIDA, we conduct ablation studies on the domain adversarial sampling (DAS), the self-adaptive rate (SAR), and the anchor buffer $\tilde{\mathcal{B}}_A$. We also explore different initialization of $p$ defined in \cref{s-adaptive-rate} to show its effect.

\textbf{Ablation on DAS and SAR.}\,\, To investigate the individual effects of DAS and SAR, we conduct ablations on each separately with all noise on both Hopper and Swimmer environments. As shown in \cref{ablation-single}-(a), removing SAR leads to a significant performance drop on Hopper and a high variance of performance on Swimmer, indicating that dynamically adjusting the proportion of $Z_I$ in $Z_{\text{mix}}$ can help assign better training samples to the policy discriminator thus improving the performance and the stability of the algorithm. Eliminating DAS results in a drastic decrease in performance, revealing that compared with uniform sampling, DAS can select embeddings with less domain information thus facilitating the training process. In the case of pure expert, w/o SAR and w/o DAS achieve comparable performance to DIDA, suggesting that our method is somehow modular-robust and the performance is desirable in the pure expert case even if SAR or DAS is absent. w/o DAS has the SAR module, while w/o SAR does not because DAS is SAR-dependent. We observe that w/o DAS performs better than w/o SAR in most cases, which also validates the effectiveness of the SAR design.

\begin{figure}[t]
    \centering
    \includegraphics[width=\columnwidth]{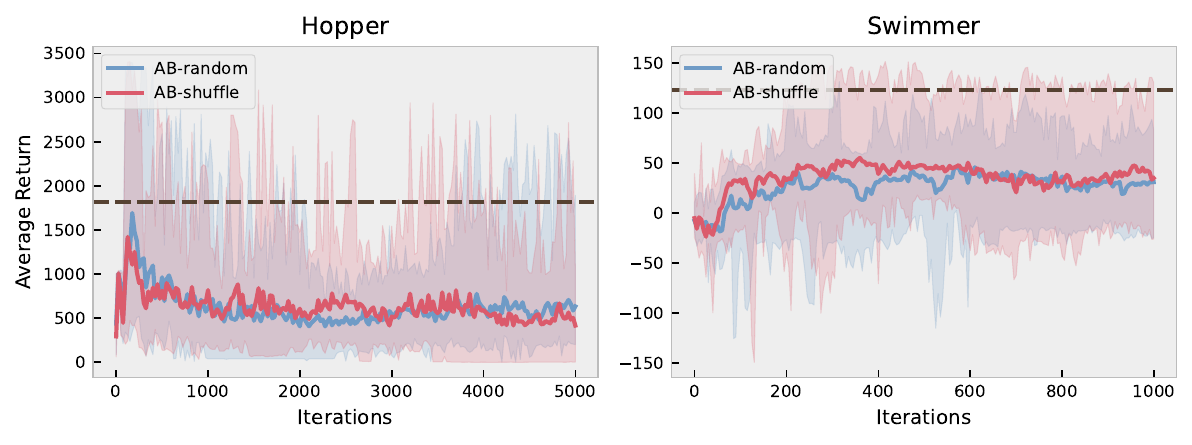}
    \caption{Performance curves of TPIL with two different anchor buffer generation methods. We conduct experiments on all noise types in two environments and present the overall average performance, where the dotted line represents the expert performance. Performance plots for each noise separately are in \cref{appendix-ablation-anchor-all-noise}.}
    \label{f-ablation-anchor}
\end{figure}

\textbf{Different generation methods of $\tilde{\mathcal{B}}_A$.}\,\, 
Considering the following two different generation methods of anchor buffer: \textit{anchor-buffer-random} collects data in the noise domain using a random policy, which is adopted in previous works; while DIDA proposes \text{anchor-buffer-shuffle}, which shuffles the noisy expert buffer randomly to simulate noisy random data. As illustrated in \cref{f-ablation-anchor}, anchor-buffer-random(AB-random) and anchor-buffer-shuffle(AB-shuffle) exhibit a negligible impact on TPIL's performance. This demonstrates that this simple data generation method is efficient. It also indirectly validates the effectiveness of other designs of DIDA including DAS and SAR. Agent-environment interaction is expensive in most real-world scenarios, making it impractical for AB-random to collect data in the expert domain. In contrast, AB-shuffle only requires a shuffling operation on the noisy expert buffer. While achieving comparable performance, our AB-shuffle method is more practical than AB-random in real-world applications.

\textbf{Different settings of SAR.}\,\, We investigate the effects of different settings of $p$ in SAR on experimental performance. We test SAR-$\frac{1}{3}$, SAR-$\frac{1}{2}$ and SAR-$\frac{2}{3}$ on three kinds of noise. As shown in \cref{ablation-single}-(b)
DIDA's default setting SAR-$\frac{2}{3}$ outstrips other settings on two noise, indicating the rationality of using the proportion of noisy data as the value of $p$. SAR-$\frac{1}{3}$ performs best with the last noise, suggesting that both $p_{acc}=\frac{1}{3}$ and $p_{acc}=\frac{2}{3}$ correspond to cases where the discriminator is unable to separate the data, which is consistent with the analysis in \cref{s-adaptive-rate}.

\subsection{Effects of different noise}
Although DIDA performs well almost on all designed tasks, it still shows variance across different noise. The additive \textit{Gaussian} noise is relatively simple, and we conduct a simple experiment using GAIL to imitate expert demonstrations under different scales of Gaussian noise. The results in \cref{f-limitation} demonstrate that the performance of GAIL remains satisfying when the noise scale is relatively small, but decreases with the noise scale increasing. 

The multiplicative noise has a greater impact on algorithm performance. We find that if the algorithm can achieve better performance on \textit{Doubly-stochastic} noise, then the performance on \textit{Shuffle} can also be guaranteed, which is consistent with the definition of the two noise in \cref{s-setting}. \textit{Normal} noise is the hardest not only for DIDA, but for other baseline methods on the two environments.

\section{Conclusion}
Our work focuses on scenarios where pure expert data or human ranking is inaccessible, thus imitator must learn from many types of noisy data. We propose Denoised Imitation learning based on Domain Adaptation (DIDA), which designs a feature encoder trying to extract task-related but domain-agnostic information from the noisy states and eliminate domain information with the help of a noise discriminator and a policy discriminator. To facilitate training, we design several efficient modules including a shuffle anchor buffer, a self-adaptive rate and a domain-adversarial sampling technique. Our experiments validate the effectiveness of our designed modules and further demonstrate the impact of various noise.

\textbf{Limitations and Future Work.}\,\, DIDA has addressed several LTI noise, and future work could consider time-varying noise and even nonlinear noise. We assume that noise can be divided into the task-irrelevant part and the task-relevant part, and focus on the task-irrelevant noise added during signal transmission. Yet task-relevant noise is inherent in the environment, which can affect the accuracy of an imitator's perception of the real world. To solve the task-relevant noise may be a promising research field in future works.

\section{Societal Impact}

This paper presents work that aims to advance the field of Machine Learning. There are many potential societal consequences of our work, none of which we feel must be specifically highlighted here.

\nocite{langley00}

\bibliography{example_paper}

\begin{thebibliography}{40}
\providecommand{\natexlab}[1]{#1}
\providecommand{\url}[1]{\texttt{#1}}
\expandafter\ifx\csname urlstyle\endcsname\relax
  \providecommand{\doi}[1]{doi: #1}\else
  \providecommand{\doi}{doi: \begingroup \urlstyle{rm}\Url}\fi

\bibitem[Belkhale et~al.(2023)Belkhale, Cui, and Sadigh]{belkhale2023hydra}
Belkhale, S., Cui, Y., and Sadigh, D.
\newblock Hydra: Hybrid robot actions for imitation learning.
\newblock In \emph{Conference on Robot Learning}, pp.\  2113--2133. PMLR, 2023.

\bibitem[Ben-David et~al.(2006)Ben-David, Blitzer, Crammer, and
  Pereira]{ben2006analysis}
Ben-David, S., Blitzer, J., Crammer, K., and Pereira, F.
\newblock Analysis of representations for domain adaptation.
\newblock \emph{Advances in neural information processing systems}, 19, 2006.

\bibitem[Brown et~al.(2019)Brown, Goo, Nagarajan, and
  Niekum]{brown2019extrapolating}
Brown, D., Goo, W., Nagarajan, P., and Niekum, S.
\newblock Extrapolating beyond suboptimal demonstrations via inverse
  reinforcement learning from observations.
\newblock In \emph{International conference on machine learning}, pp.\
  783--792. PMLR, 2019.

\bibitem[Brown et~al.(2020)Brown, Goo, and Niekum]{brown2020better}
Brown, D.~S., Goo, W., and Niekum, S.
\newblock Better-than-demonstrator imitation learning via automatically-ranked
  demonstrations.
\newblock In \emph{Conference on robot learning}, pp.\  330--359. PMLR, 2020.

\bibitem[Burchfiel et~al.(2016)Burchfiel, Tomasi, and
  Parr]{burchfiel2016distance}
Burchfiel, B., Tomasi, C., and Parr, R.
\newblock Distance minimization for reward learning from scored trajectories.
\newblock In \emph{Proceedings of the AAAI Conference on Artificial
  Intelligence}, volume~30, 2016.

\bibitem[Cetin \& Celiktutan(2021)Cetin and Celiktutan]{cetin2021domain}
Cetin, E. and Celiktutan, O.
\newblock Domain-robust visual imitation learning with mutual information
  constraints.
\newblock \emph{arXiv preprint arXiv:2103.05079}, 2021.

\bibitem[Chae et~al.(2022)Chae, Han, Jung, Cho, Choi, and Sung]{chae2022robust}
Chae, J., Han, S., Jung, W., Cho, M., Choi, S., and Sung, Y.
\newblock Robust imitation learning against variations in environment dynamics.
\newblock In \emph{International Conference on Machine Learning}, pp.\
  2828--2852. PMLR, 2022.

\bibitem[Chen et~al.(2021)Chen, Paleja, and Gombolay]{chen2021learning}
Chen, L., Paleja, R., and Gombolay, M.
\newblock Learning from suboptimal demonstration via self-supervised reward
  regression.
\newblock In \emph{Conference on robot learning}, pp.\  1262--1277. PMLR, 2021.

\bibitem[Chen et~al.(2023)Chen, Bai, Poor, and Wang]{chen2023efficient}
Chen, M., Bai, Y., Poor, H.~V., and Wang, M.
\newblock Efficient rl with impaired observability: Learning to act with
  delayed and missing state observations.
\newblock \emph{arXiv preprint arXiv:2306.01243}, 2023.

\bibitem[Davis et~al.(2023)Davis, Katz, Gentili, and
  Reggia]{davis2023neuroceril}
Davis, G.~P., Katz, G.~E., Gentili, R.~J., and Reggia, J.~A.
\newblock Neuroceril: Robotic imitation learning via hierarchical cause-effect
  reasoning in programmable attractor neural networks.
\newblock \emph{International Journal of Social Robotics}, pp.\  1--19, 2023.

\bibitem[Duan et~al.(2023)Duan, Batzianoulis, Camoriano, Rosasco, Pucci, and
  Billard]{duan2023structured}
Duan, A., Batzianoulis, I., Camoriano, R., Rosasco, L., Pucci, D., and Billard,
  A.
\newblock A structured prediction approach for robot imitation learning.
\newblock \emph{The International Journal of Robotics Research}, pp.\
  02783649231204656, 2023.

\bibitem[Fickinger et~al.(2021)Fickinger, Cohen, Russell, and
  Amos]{fickinger2021cross}
Fickinger, A., Cohen, S., Russell, S., and Amos, B.
\newblock Cross-domain imitation learning via optimal transport.
\newblock \emph{arXiv preprint arXiv:2110.03684}, 2021.

\bibitem[Gandhi et~al.(2023)Gandhi, Karamcheti, Liao, and
  Sadigh]{gandhi2023eliciting}
Gandhi, K., Karamcheti, S., Liao, M., and Sadigh, D.
\newblock Eliciting compatible demonstrations for multi-human imitation
  learning.
\newblock In \emph{Conference on Robot Learning}, pp.\  1981--1991. PMLR, 2023.

\bibitem[Ganin et~al.(2016)Ganin, Ustinova, Ajakan, Germain, Larochelle,
  Laviolette, March, and Lempitsky]{ganin2016domain}
Ganin, Y., Ustinova, E., Ajakan, H., Germain, P., Larochelle, H., Laviolette,
  F., March, M., and Lempitsky, V.
\newblock Domain-adversarial training of neural networks.
\newblock \emph{Journal of machine learning research}, 17\penalty0
  (59):\penalty0 1--35, 2016.

\bibitem[Ghasemipour et~al.(2020)Ghasemipour, Zemel, and
  Gu]{ghasemipour2020divergence}
Ghasemipour, S. K.~S., Zemel, R., and Gu, S.
\newblock A divergence minimization perspective on imitation learning methods.
\newblock In \emph{Conference on Robot Learning}, pp.\  1259--1277. PMLR, 2020.

\bibitem[Goodfellow et~al.(2014)Goodfellow, Pouget-Abadie, Mirza, Xu,
  Warde-Farley, Ozair, Courville, and Bengio]{goodfellow2014generative}
Goodfellow, I., Pouget-Abadie, J., Mirza, M., Xu, B., Warde-Farley, D., Ozair,
  S., Courville, A., and Bengio, Y.
\newblock Generative adversarial nets.
\newblock \emph{Advances in neural information processing systems}, 27, 2014.

\bibitem[Gu \& Dao(2023)Gu and Dao]{gu2023mamba}
Gu, A. and Dao, T.
\newblock Mamba: Linear-time sequence modeling with selective state spaces.
\newblock \emph{arXiv preprint arXiv:2312.00752}, 2023.

\bibitem[Haliem et~al.(2021)Haliem, Bonjour, Alsalem, Thomas, Li, Aggarwal,
  Bhargava, and Kejriwal]{haliem2021learning}
Haliem, M., Bonjour, T., Alsalem, A., Thomas, S., Li, H., Aggarwal, V.,
  Bhargava, B., and Kejriwal, M.
\newblock Learning monopoly gameplay: A hybrid model-free deep reinforcement
  learning and imitation learning approach.
\newblock \emph{arXiv preprint ArXiv:2103.00683}, 2021.

\bibitem[Ho \& Ermon(2016)Ho and Ermon]{ho2016generative}
Ho, J. and Ermon, S.
\newblock Generative adversarial imitation learning.
\newblock \emph{Advances in neural information processing systems}, 29, 2016.

\bibitem[Hussein et~al.(2017)Hussein, Gaber, Elyan, and
  Jayne]{hussein2017imitation}
Hussein, A., Gaber, M.~M., Elyan, E., and Jayne, C.
\newblock Imitation learning: A survey of learning methods.
\newblock \emph{ACM Computing Surveys (CSUR)}, 50\penalty0 (2):\penalty0 1--35,
  2017.

\bibitem[Ke et~al.(2021)Ke, Choudhury, Barnes, Sun, Lee, and
  Srinivasa]{ke2021imitation}
Ke, L., Choudhury, S., Barnes, M., Sun, W., Lee, G., and Srinivasa, S.
\newblock Imitation learning as f-divergence minimization.
\newblock In \emph{Algorithmic Foundations of Robotics XIV: Proceedings of the
  Fourteenth Workshop on the Algorithmic Foundations of Robotics 14}, pp.\
  313--329. Springer, 2021.

\bibitem[Langley(2000)]{langley00}
Langley, P.
\newblock Crafting papers on machine learning.
\newblock In Langley, P. (ed.), \emph{Proceedings of the 17th International
  Conference on Machine Learning (ICML 2000)}, pp.\  1207--1216, Stanford, CA,
  2000. Morgan Kaufmann.

\bibitem[Laskey et~al.(2017)Laskey, Lee, Fox, Dragan, and
  Goldberg]{laskey2017dart}
Laskey, M., Lee, J., Fox, R., Dragan, A., and Goldberg, K.
\newblock Dart: Noise injection for robust imitation learning.
\newblock In \emph{Conference on robot learning}, pp.\  143--156. PMLR, 2017.

\bibitem[Liu et~al.(2019)Liu, Ling, Mu, and Su]{liu2019state}
Liu, F., Ling, Z., Mu, T., and Su, H.
\newblock State alignment-based imitation learning.
\newblock \emph{arXiv preprint arXiv:1911.10947}, 2019.

\bibitem[Osa et~al.(2018)Osa, Pajarinen, Neumann, Bagnell, Abbeel, Peters,
  et~al.]{osa2018algorithmic}
Osa, T., Pajarinen, J., Neumann, G., Bagnell, J.~A., Abbeel, P., Peters, J.,
  et~al.
\newblock An algorithmic perspective on imitation learning.
\newblock \emph{Foundations and Trends{\textregistered} in Robotics},
  7\penalty0 (1-2):\penalty0 1--179, 2018.

\bibitem[Peng \& Lee(2023)Peng and Lee]{peng2023valuation}
Peng, Y.-L. and Lee, W.-P.
\newblock Valuation of stocks by integrating discounted cash flow with
  imitation learning and guided policy.
\newblock \emph{IEEE Transactions on Automation Science and Engineering}, 2023.

\bibitem[Ross \& Bagnell(2010)Ross and Bagnell]{ross2010efficient}
Ross, S. and Bagnell, D.
\newblock Efficient reductions for imitation learning.
\newblock In \emph{Proceedings of the thirteenth international conference on
  artificial intelligence and statistics}, pp.\  661--668. JMLR Workshop and
  Conference Proceedings, 2010.

\bibitem[Schaal(1999)]{schaal1999imitation}
Schaal, S.
\newblock Is imitation learning the route to humanoid robots?
\newblock \emph{Trends in cognitive sciences}, 3\penalty0 (6):\penalty0
  233--242, 1999.

\bibitem[Schulman et~al.(2017)Schulman, Wolski, Dhariwal, Radford, and
  Klimov]{schulman2017proximal}
Schulman, J., Wolski, F., Dhariwal, P., Radford, A., and Klimov, O.
\newblock Proximal policy optimization algorithms.
\newblock \emph{arXiv preprint arXiv:1707.06347}, 2017.

\bibitem[Sharma et~al.(2019)Sharma, Pathak, and Gupta]{sharma2019third}
Sharma, P., Pathak, D., and Gupta, A.
\newblock Third-person visual imitation learning via decoupled hierarchical
  controller.
\newblock \emph{Advances in Neural Information Processing Systems}, 32, 2019.

\bibitem[Shi et~al.(2023)Shi, Xu, Fang, and Chen]{shi2023self}
Shi, Z., Xu, Y., Fang, M., and Chen, L.
\newblock Self-imitation learning for action generation in text-based games.
\newblock In \emph{Proceedings of the 17th Conference of the European Chapter
  of the Association for Computational Linguistics}, pp.\  703--726, 2023.

\bibitem[Stadie et~al.(2017)Stadie, Abbeel, and Sutskever]{stadie2017third}
Stadie, B.~C., Abbeel, P., and Sutskever, I.
\newblock Third-person imitation learning.
\newblock \emph{arXiv preprint arXiv:1703.01703}, 2017.

\bibitem[Tangkaratt et~al.(2020)Tangkaratt, Charoenphakdee, and
  Sugiyama]{tangkaratt2020robust}
Tangkaratt, V., Charoenphakdee, N., and Sugiyama, M.
\newblock Robust imitation learning from noisy demonstrations.
\newblock \emph{arXiv preprint arXiv:2010.10181}, 2020.

\bibitem[Todorov et~al.(2012)Todorov, Erez, and Tassa]{todorov2012mujoco}
Todorov, E., Erez, T., and Tassa, Y.
\newblock Mujoco: A physics engine for model-based control.
\newblock In \emph{2012 IEEE/RSJ international conference on intelligent robots
  and systems}, pp.\  5026--5033. IEEE, 2012.

\bibitem[Van~der Maaten \& Hinton(2008)Van~der Maaten and
  Hinton]{van2008visualizing}
Van~der Maaten, L. and Hinton, G.
\newblock Visualizing data using t-sne.
\newblock \emph{Journal of machine learning research}, 9\penalty0 (11), 2008.

\bibitem[Wang et~al.(2022)Wang, Zhuang, Wang, and Zhao]{wang2022adversarially}
Wang, J., Zhuang, Z., Wang, Y., and Zhao, H.
\newblock Adversarially robust imitation learning.
\newblock In \emph{Conference on Robot Learning}, pp.\  320--331. PMLR, 2022.

\bibitem[Wu et~al.(2019)Wu, Charoenphakdee, Bao, Tangkaratt, and
  Sugiyama]{wu2019imitation}
Wu, Y.-H., Charoenphakdee, N., Bao, H., Tangkaratt, V., and Sugiyama, M.
\newblock Imitation learning from imperfect demonstration.
\newblock In \emph{International Conference on Machine Learning}, pp.\
  6818--6827. PMLR, 2019.

\bibitem[Yuan et~al.(2023)Yuan, Li, Heng, Zhang, and Wang]{yuan2023good}
Yuan, Y., Li, X., Heng, Y., Zhang, L., and Wang, M.
\newblock Good better best: Self-motivated imitation learning for noisy
  demonstrations.
\newblock \emph{arXiv preprint arXiv:2310.15815}, 2023.

\bibitem[Zhang et~al.(2021)Zhang, Cao, Sadigh, and Sui]{zhang2021confidence}
Zhang, S., Cao, Z., Sadigh, D., and Sui, Y.
\newblock Confidence-aware imitation learning from demonstrations with varying
  optimality.
\newblock \emph{Advances in Neural Information Processing Systems},
  34:\penalty0 12340--12350, 2021.

\bibitem[Zhang et~al.(2023)Zhang, Kang, Lee, Tomlin, Levine, Tu, and
  Matni]{zhang2023multi}
Zhang, T.~T., Kang, K., Lee, B.~D., Tomlin, C., Levine, S., Tu, S., and Matni,
  N.
\newblock Multi-task imitation learning for linear dynamical systems.
\newblock In \emph{Learning for Dynamics and Control Conference}, pp.\
  586--599. PMLR, 2023.

\end{thebibliography}
\bibliographystyle{icml2024}


\newpage
\appendix
\onecolumn
\section{Algorithm}
\begin{algorithm}
    \renewcommand{\algorithmicrequire}{\textbf{Input:}}
    \renewcommand{\algorithmicensure}{\textbf{Output:}}
    \caption{DIDA: Denoised Imitation Learning based on Domain Adaptation}
    \label{alg1}
    \begin{algorithmic}[1]
    \REQUIRE Expert buffer $\mathcal{B}_E$, empty buffers $\{\tilde{\mathcal{B}}_E,\tilde{\mathcal{B}}_A,\mathcal{B}_I\}$, noise operator $f_{\text{noise}}$ and initial domain weight $\lambda_0$
    \STATE Randomly initialize feature encoder $G_f$, policy net $\pi$, value net $V_{\theta}$ and discriminators $\left\{D_n,D_p\right\}$ 
    \STATE $\tilde{\mathcal{B}}_E = f_{\text{noise}}(\mathcal{B}_E)$, and randomly shuffle $\tilde{\mathcal{B}}_E$ to get $\tilde{\mathcal{B}}_A$
    
    \FOR{epoch $i=1:M$}
    \STATE Sample a batch of indexes from $|\tilde{\mathcal{B}}_E|$ and extract corresponding data in $\tilde{\mathcal{B}}_E$ and $|\tilde{\mathcal{B}}_A|$ to get $\tilde{S}_E$ and $\tilde{S}_A$
    \STATE\textcolor{red}{// Collect training data through interacting with environment}
    \STATE $\mathcal{B}_I=\{s_t,a_t,r_t,s_{t+1}\}_{t=1}^N$, where $a_t\sim\pi^i(s_t)$, $h_t=\text{concat}(G_f(s_t),G_f(s_{t+1}))$ and $r_t=D_p(h_t)$
    \STATE Extract states from $\mathcal{B}_I$ to get $S_I=\{s_t,s_{t+1}\}_{t=1}^N$
    \STATE Compute values $V_i=\{V_{\theta}^i(s_t)\}_{t=1}^N$, returns $G_i=\{\sum_{j=1}^t\gamma^j r_j\}_{t=1}^N$ and advantages $A_i=\{a_t\}_{t=1}^N$
    \STATE Get embeddings $(Z_I,\tilde{Z}_E,\tilde{Z}_A)$ by feeding $(S_I,\tilde{S}_E,\tilde{S}_A)$ into $G_f$

    \STATE\textcolor{red}{// Update noise and policy discriminators}
    \STATE Compute loss between predictions $D_n(Z_I,\tilde{Z}_E,\tilde{Z}_A)$ and labels (0 for $Z_I$, 1 for $\tilde{Z}_E$ and $\tilde{Z}_A$)
    \STATE Compute the gradients $g_n$ and perform backpropagation to update $D_n$
    \STATE Calculate the adaptive rate $\alpha$ using the method described in \cref{s-adaptive-rate}
    \STATE Compute the domain weight $\lambda=\lambda_0(2 / (1 + e^{-10q}) - 1)$, where $q=\frac{i}{M}$
    \STATE $Z_{\text{mix}}=\text{DAS}(\tilde{Z}_A,Z_I,\alpha)$
    \STATE Compute loss between predictions $D_p(\tilde{Z}_E,Z_{\text{mix}})$ and labels (0 for $Z_{\text{mix}}$ and 1 for $\tilde{Z}_E$)
    \STATE Compute the gradients $g_p$ and perform backpropagation to update $D_p$

    \STATE\textcolor{red}{// Update feature encoder}
    \STATE Use $g_p-\lambda g_n$ to update $G_f$, where negative sign comes from reverse gradient layer $R$

    \STATE\textcolor{red}{// Perform PPO update}
    \STATE $\pi^{i+1},V_{\theta}^{i+1} = \text{ppo\_update}(\pi^i,V_{\theta}^i,\mathcal{B}_I,G_i,A_i)$
    \ENDFOR
\end{algorithmic}  
\end{algorithm}

\section{Proof}
\begin{figure*}[h]
    \centering
    \includegraphics[width=0.5\textwidth]{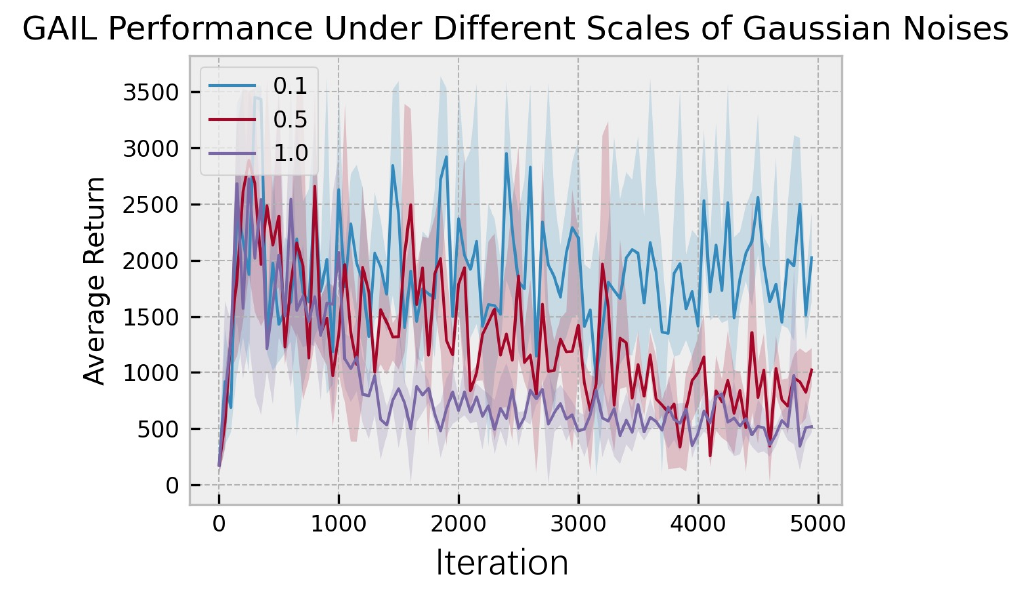}
    \caption{Performance of GAIL on different scales of \textbf{Gaussian} noise.} 
    \label{f-limitation}
\end{figure*}

\begin{proof}[Proof of Proposition 4.1]
    \label{proof}
    GAIL \cite{ho2016generative} has proved that a policy and its occupancy measure are one-to-one, and they can be converted into each other under the same MDP assumption. GAIL indirectly imitates an expert policy by matching the occupancy measures \cite{ho2016generative}. When the discriminator is unable to distinguish between the two distributions and the agent achieves a high return in evaluation, we believe that the agent has learned a policy that is close to the expert policy. However, the LND problem involves two different MDPs $\mathcal{M}$ and $\mathcal{M}_n$. We believe that, under the influence of LTI noise, there exists a certain correspondence between the expert policy $\pi_e$ (derived from dataset $\mathcal{B}_E$) in the original MDP $\mathcal{M}$ and the noisy expert policy $\tilde{\pi}_e$ (derived from dataset $\tilde{\mathcal{B}}_E$) in the noisy MDP $\mathcal{M}_n$. The properties of LTI noise imply $\pi_e(s|a)=\tilde{\pi}_e(a|\tilde{s})$. We try to prove the following equation:
    \begin{align}
        \label{19}
        P(s_t=s|\pi_e) = P(s_t=\tilde{s}|\tilde{\pi}_e)
    \end{align}
    Let's prove it by mathematical induction, first considering the case $t=0$, easily we can get:
    \begin{align}
        \label{20}
        P(s_0=s|\pi_e) = p_0(s) = p_0^n(\tilde{s}) = P(s_0=\tilde{s}|\tilde{\pi}_e)
    \end{align}
    where $\tilde{s}=f_{\text{noise}}(s)$, and \cref{20} holds for all $s\in\mathcal{S}$. We assume that the equation holds for $t=n-1$, let us consider the case of $t=n$:
    \begin{align}
        P(s_n=s|\pi_e) &= \sum_{s'}P(s_{n-1}=s'|\pi_e)\sum_a\pi_e(a|s')\mathcal{P}(s|s',a) \\
        &= \sum_{s'}P(s_{n-1}=s'|\pi_e)\sum_a\tilde{\pi}_e(a|\tilde{s}')\mathcal{P}_n(\tilde{s}|\tilde{s}',a) \\
        &= \sum_{\tilde{s}'}P(s_{n-1}=\tilde{s}'|\tilde{\pi}_e)\sum_a\tilde{\pi}_e(a|\tilde{s}')\mathcal{P}_n(\tilde{s}|\tilde{s}',a) \\
        &= P(s_n=\tilde{s}|\tilde{\pi}_e)
    \end{align}
    Thus we finish the proof of \cref{19}. Considering the definitions of $\rho_e$ and $\tilde{\rho}_e$, we have:
    \begin{align}
        \rho_{\pi_e}(s,a) &= (1-\gamma)\pi_e(a|s)\sum_{t=0}^{\infty}\gamma^tP(s_t=s|\pi_e) \\
        &= (1-\gamma)\tilde{\pi}_e(a|\tilde{s})\sum_{t=0}^{\infty}\gamma^tP(s_t=\tilde{s}|\tilde{\pi}_e) \\
        \label{18}
        &= \rho_{\tilde{\pi}_e}(\tilde{s},a)
    \end{align}
    LTI noise serves as a one-to-one mapping between $s$ and $\tilde{s}$, so $\pi_e(a|s)=\tilde{\pi}_e(s|\tilde{s})$. Combining this with \cref{19}, we can get \cref{18}. That is, although the occupation measures of the two policies may be equal $\rho_e(s,a)=\tilde{\rho}_e(f_{\text{noise}}(s),a)$, they are actually two completely different policies. Directly using dataset $\tilde{\mathcal{B}}_E$ and dataset $\mathcal{B}_A$ for GAIL learning would lead to training failure due to the gap between MDPs. 
    Under the same MDP, $\rho(s,a)=\rho_e(s,a)$ and $\pi=pi_e$ both are necessary and sufficient conditions for each other for the training of GAIL. Therefore, only if $\rho_e(\tilde{s},a)$ is as close to $\rho_e(s,a)$ as possible can GAIL solve the LND problem, and the difference between the two measures the gap between MDPs. We now analyze the feasibility of this condition:
    \begin{align}
        \label{11}
        \rho_{\pi_e}(\tilde{s},a) &= (1-\gamma)\pi_e(a|\tilde{s})\sum_{t=0}^{\infty}\gamma^tP(s_t=\tilde{s}|\pi_e) \\
        \label{12}
        &= (1-\gamma)\pi_e(a|s)\sum_{t=0}^{\infty}\gamma^tP(s_t=s|\pi_e)\frac{\pi_e(a|\tilde{s})}{\pi_e(a|s)}|A^{-1}| \\
        \label{13}
        &= \rho_{\pi_e}(s,a)\frac{\pi_e(a|\tilde{s})}{\pi_e(a|s)}|A^{-1}| \\
        \label{14}
        &=^* \rho_{\pi_e}(s,a)\frac{\pi_e(a|\tilde{s})}{\pi_e(a|s)} \\
        \label{15}
        &=^* \rho_{\pi_e}(s,a) 
    \end{align}
    \cref{11} follows from the definition of occupancy measure, and \cref{12} is based on the probability density transformation formula $P(\tilde{s})=P(f_{\text{noise}}^{-1}(\tilde{s}))|(f_{\text{noise}}^{-1})'(\tilde{s})|$. Only under the \textit{Gaussian} noise or the special case of the \textit{Doubly-stochastic} noise (the identity matrix is a type of doubly-stochastic matrix), $A$ satisfies the requirements of an identity matrix, and \cref{14} holds. The validity of \cref{15} requires either of the following two conditions:
    \begin{itemize}
        \item 1) Expert $\pi_e$ is robust to noise. For example, human experts can see the essence of the phenomenon when facing problems, and then make correct decisions.
        \item 2) The noise added is relatively small compared to the original state, such as Gaussian noise with small $\mu$ and $\sigma$. Then the agent can take similar action for the noisy state $\tilde{s}$ and original state $s$.
    \end{itemize}
    Only if the noise is small enough to satisfy the second condition and the \cref{15} holds, GAIL can solve this case. Otherwise, the fraction term $\frac{\pi_e(a|\tilde{s})}{\pi_e(a|s)}$ cannot be ignored, then $\rho_{\pi_e}(\tilde{s},a) \neq \rho_{\pi_e}(s,a)$ and GAIL cannot learn well.
\end{proof}
In the above derivation, we found that two factors affect the difficulty of solving LND problem: the first is $|A^{-1}|$, which is determined by the nature of the problem itself; the second is $\frac{\pi_e(a|\tilde{s})}{\pi_e(a|s)}$, which can characterize the ability of the expert policy to resist noise. If we can estimate the value of $\frac{\pi_e(a|\tilde{s})}{\pi_e(a|s)}$ through importance sampling or off-policy techniques, it may help solve the LND problem. To verify the correctness of \cref{pro1}, we evaluated the performance of GAIL on different scales of \textit{Gaussian} noise. We fixed the mean of the noise to $0$ and set the standard deviation to $0.1$, $0.5$, and $1.0$ respectively. As can be seen from \cref{f-limitation}, the policy can resist noise with the scale of $0.1$, so it satisfies $\pi_e(a|s)\approx\pi_e(s|\tilde{s})$ and the final performance can be guaranteed. As the noise scale increases, the performance of GAIL gradually deteriorates, which is consistent with the result of \cref{pro1}.

\section{Additional results}
\subsection{Performance curves of \cref{s-baseline}}
\begin{figure}[H]
    \centering
    \includegraphics[width=\textwidth]{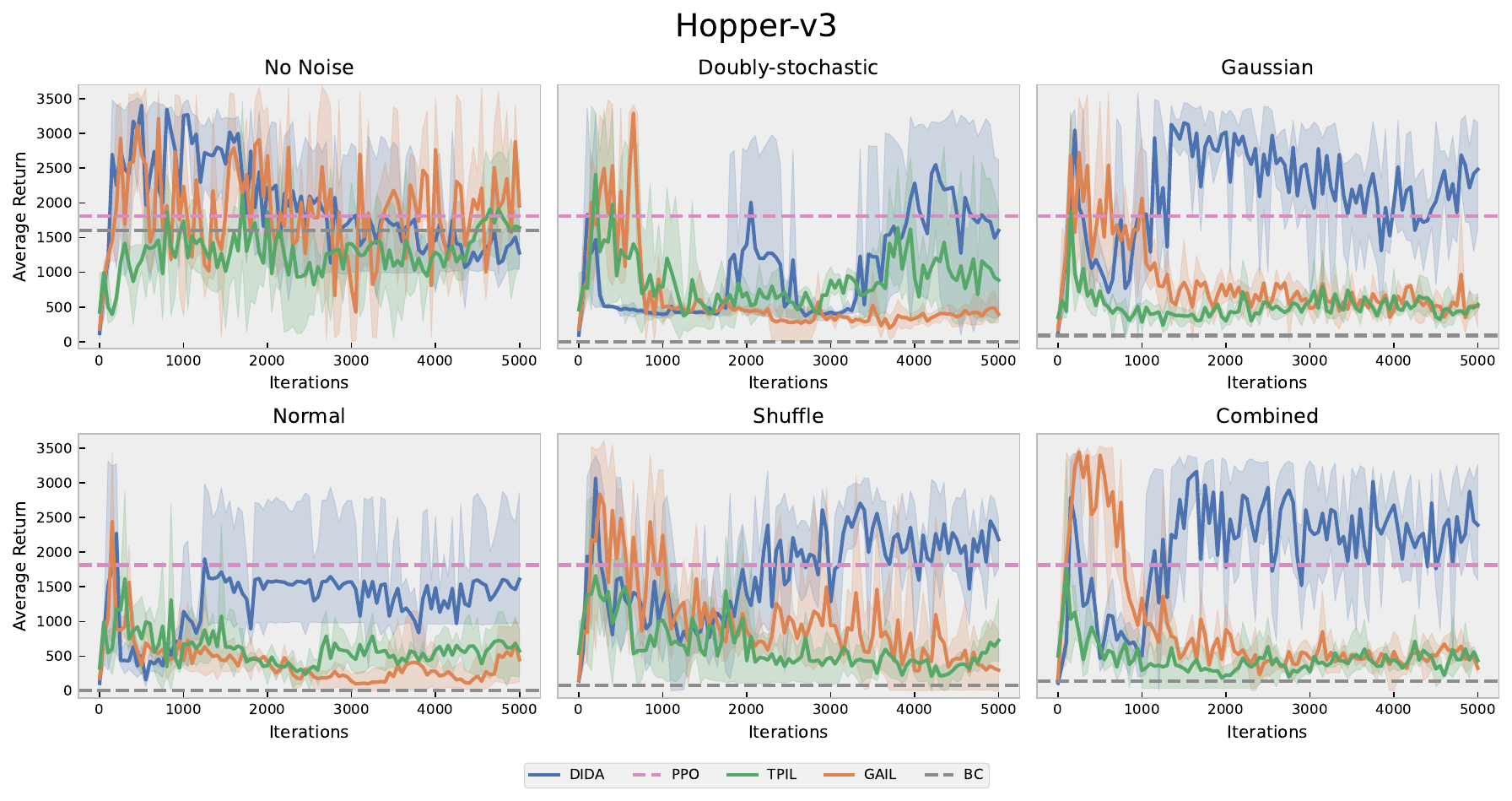}
    \caption{Performance curves of \cref{s-baseline} on Hopper.} 
    \label{f-hopper-curves}
\end{figure}
\begin{figure}[H]
    \centering
    \includegraphics[width=\textwidth]{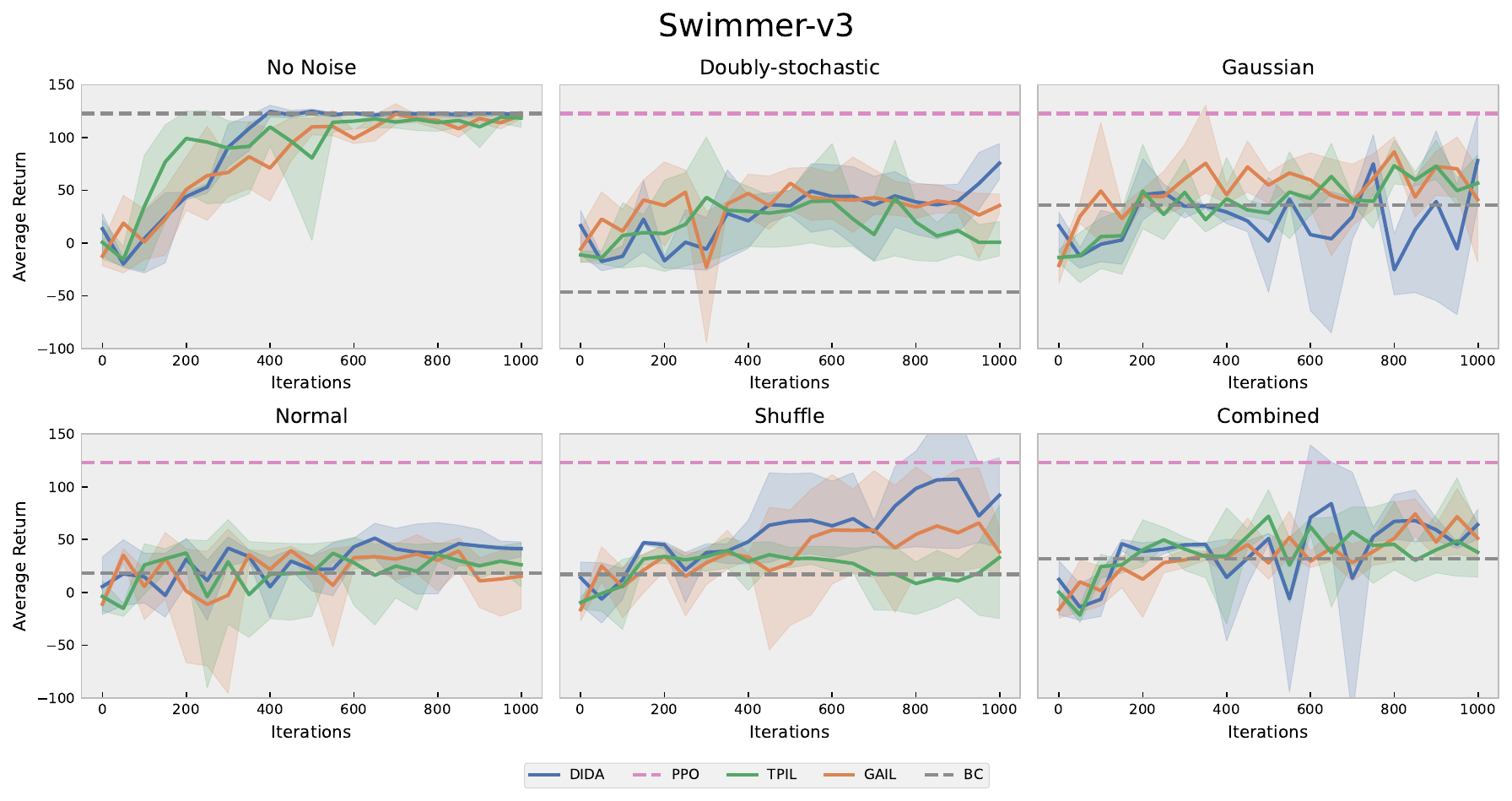}
    \caption{Performance curves of \cref{s-baseline} on Swimmer.} 
    \label{f-swimmer-curves}
\end{figure}
We plot the performance curves of DIDA, TPIL, and GAIL over time, with the hopper environment in Figure 1 and the swimmer environment in Figure 2. The pink dotted line represents the return of the expert policy trained with PPO, and the gray dotted line shows the performance of BC. The SAIL curve is not included in the figure because its performance is too poor to be plotted on the same figure with other methods.

\subsection{The full performance figures of \cref{f-ablation-anchor}}
\label{appendix-ablation-anchor-all-noise}
\begin{figure}[H]
    \centering
    \includegraphics[width=0.8\textwidth]{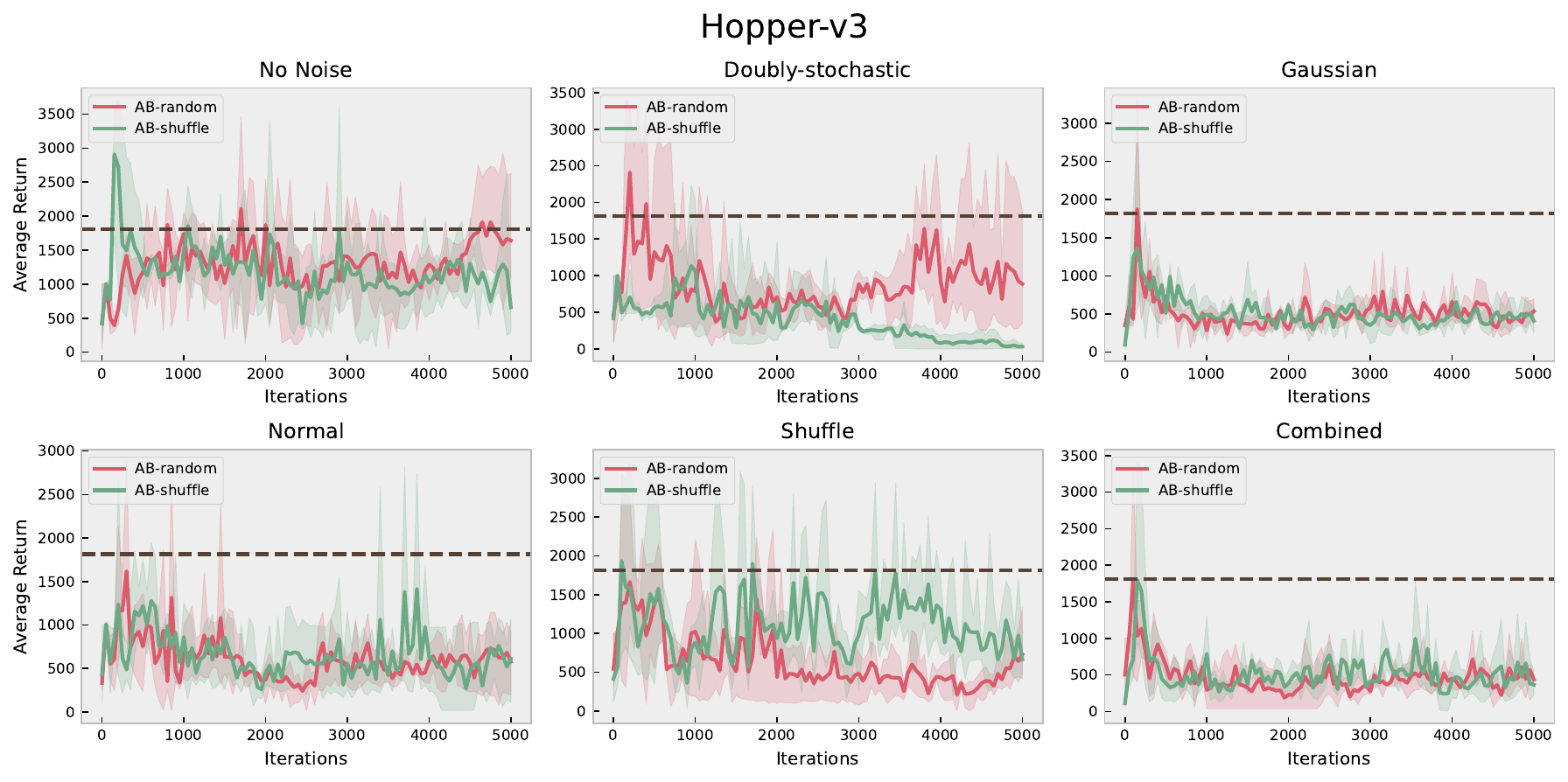}
    \caption{Performance comparison of TPIL methods with anchor-buffer-random and anchor-buffer-shuffle on multiple noisy datasets in the Hopper environment.}
    \label{exp-ablation-anchor-generation-hopper}
\end{figure}

\begin{figure}[H]
    \centering
    \includegraphics[width=0.8\textwidth]{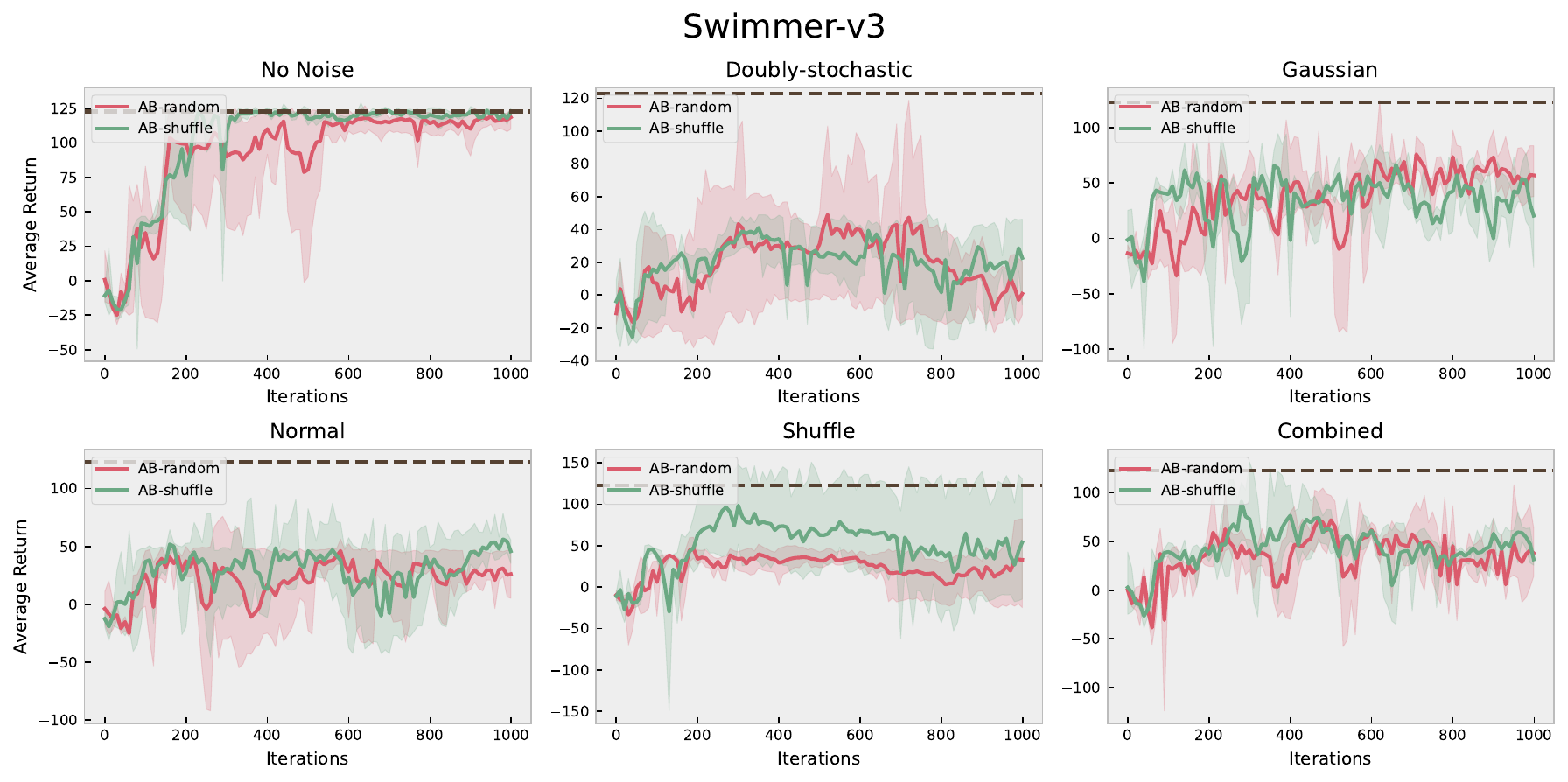}
    \caption{Performance comparison of TPIL methods with anchor-buffer-random and anchor-buffer-shuffle on multiple noisy datasets in the Swimmer environment.}
    \label{exp-ablation-anchor-generation-swimmer}
\end{figure}

We evaluated the performance difference of anchor-buffer-random and anchor-buffer-shuffle on multiple noisy datasets in the Hopper and Swimmer environments, with TPIL as the baseline method. As shown in \cref{exp-ablation-anchor-generation-hopper} and \cref{exp-ablation-anchor-generation-swimmer}, the two anchor buffer generation methods do not have a significant impact on performance. In most tasks, anchor-buffer-random performs slightly better than anchor-buffer-shuffle with smaller variance, because anchor-buffer-random obtains the ideal anchor dataset; while anchor-buffer-shuffle tries to imitate the outputs of anchor-buffer-random. However, as mentioned in \cref{exp-ablation}, anchor-buffer-shuffle is more reasonable and practical, and our method DIDA with anchor-buffer-shuffle eventually achieves better performance than other baseline methods, which further demonstrates the superiority of DIDA.

\subsection{Rendered trajectories}
\begin{figure}[H]
    \centering
    \includegraphics[width=\textwidth]{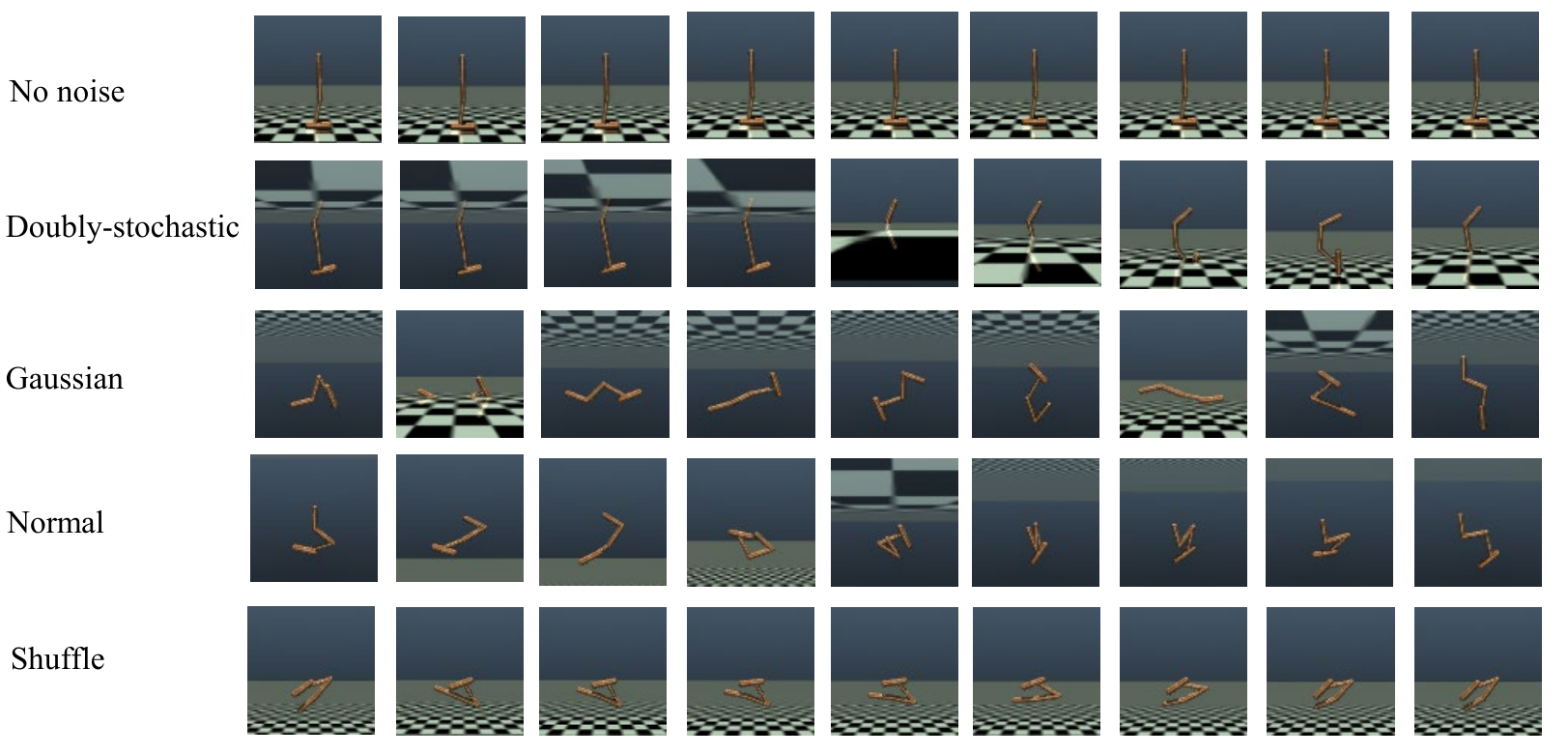}
    \caption{The rendered trajectories from five types of noisy states.} 
\end{figure}

\subsection{The t-SNE plots on latent embeddings of \cref{f-distribution-change}}
\label{appendix-confusion-tsne}
\begin{figure}[H]
    \centering
    \includegraphics[width=0.8\textwidth]{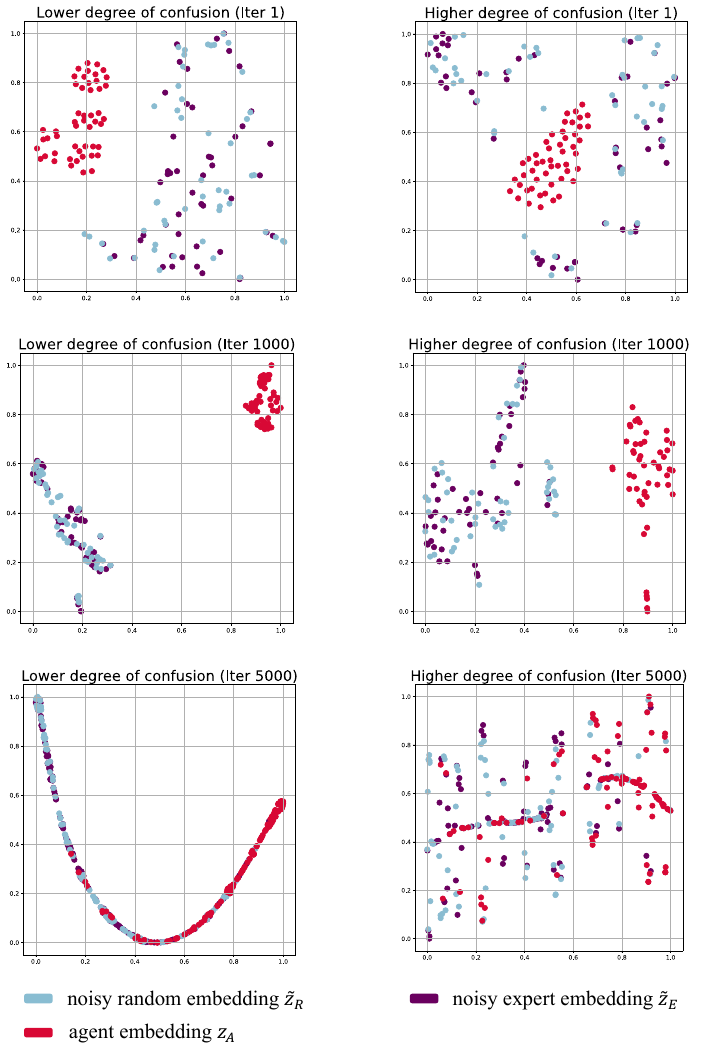}
    \caption{The t-SNE plots on latent embeddings of iter-1, iter-1000 and iter-5000, corresponding to the three rows from top to bottom. For each row, the two t-SNE plots corresponding to sample's embedding in $Z_A$, $\tilde{Z}_R$, and $\tilde{Z}_E$. The left plot corresponds to the 100 samples with the smallest degree of confusion while the right one corresponds to the 100 samples with the largest degree of confusion.} 
    \label{confusion-tsne}
\end{figure}

The first row of \cref{confusion-tsne} shows the results of the untrained model. The two clusters are close to each other because the randomly initialized feature encoder cannot extract embeddings that strongly correlated with the categories. The noise discriminator will likely classify data from the three different datasets into the same category. The second row of \cref{confusion-tsne} shows the results of the model after training for 1000 iterations. The clusters in the left and right figures are far apart, indicating that the embeddings at this time contain plenty of domain information. The third row of \cref{confusion-tsne} shows the results of the model after training for 5000 iterations. Most of the domain information in the embeddings has been eliminated at this time, and the points from different datasets in the t-SNE plot are mixed together and difficult to distinguish.

\subsection{The full t-SNE plot on embeddings of different noise}

\begin{figure}[H]
    \centering
    \includegraphics[width=0.6\textwidth]{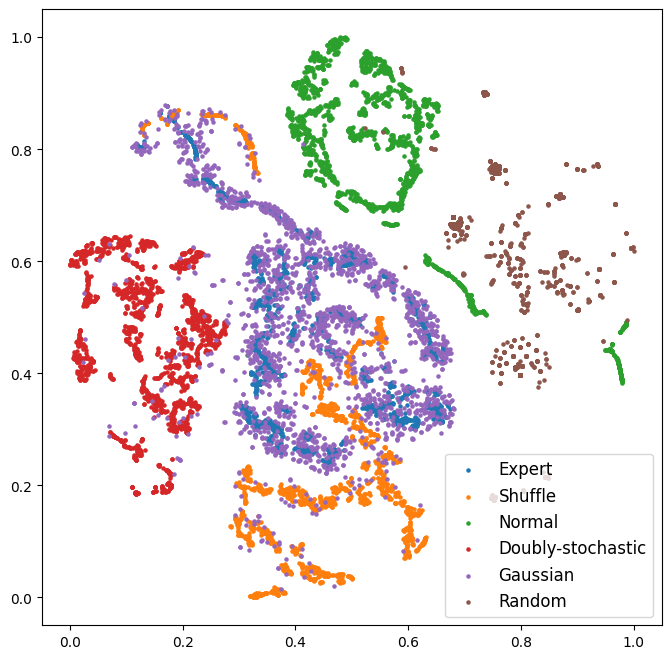}
    \caption{The t-SNE plot of the states from the random policy dataset, the expert policy dataset, as well as various noisy datasets. Different colors represent different categories. We can observe the characteristics of different noise, e.g., Gaussian noise is concentrated around the expert data, and random noise is the most dispersed. The distance between points in the t-SNE plot reflects the similarity of the data distributions.} 
    \label{f-full-tsne}
\end{figure}

\section{Implementation Details}
\subsection{Detailed Description about Baselines}\label{baselines}
\textbf{BC}\,\, Behavior cloning is a type of imitation learning that aims to learn a policy by observing the behavior of an expert. Specifically, it learns a mapping from states to actions from expert demonstration data. This mapping can be a supervised learning model, such as a regression or classification model, or a reinforcement learning model.

\textbf{PPO} \cite{schulman2017proximal}\,\, Proximal policy optimization (PPO) combines the benefits of trust region policy optimization (TRPO) with simpler implementation, better sample complexity, and improved performance in benchmark tasks. It proposes a novel objective function that allows for multiple epochs of minibatch updates, resulting in better optimization. PPO is tested on various tasks, including simulated robotic locomotion and Atari game playing, and outperforms other online policy gradient methods. The algorithm strikes a favorable balance between sample complexity, simplicity, and wall-time.

\textbf{GWIL} \cite{fickinger2021cross}\,\, Gromov-Wasserstein Imitation Learning (GWIL) is designed for cross-domain imitation learning and it utilizes the Gromov-Wasserstein distance to align and compare states between the different spaces of the expert and imitation agents. By minimizing the Gromov-Wasserstein distance, GWIL aims to find an optimal coupling between the states of the expert and imitation agents. GWIL has been demonstrated to be effective in non-trivial continuous control domains, ranging from simple rigid transformations to arbitrary transformations of the state-action space.

\textbf{SAIL} \cite{liu2019state}\,\, SAIL focuses on state alignment-based imitation learning, where the imitator and the expert have different dynamics models, and it aims to train the imitator to follow the state sequences in expert demonstrations as closely as possible. The state alignment is achieved by combining both local and global perspectives, which are integrated into a reinforcement learning framework using a regularized policy update objective. The methodology of SAIL also involves maximizing the curriculum reward, which is equivalent to optimizing the state visitation distributions for the Wasserstein distance. It's worth mentioning that the official code that the authors of SAIL open-sourced is still under pre-release version, so the official implementation may not match the reported performance in the original paper. We run 3 seeds for each designed task and report SAIL's best performance.

\textbf{GAIL} \cite{ho2016generative}\,\, This work proposes a new general framework called GAIL for directly extracting a policy from data, as if it were obtained by reinforcement learning following inverse reinforcement learning. The framework draws an analogy between imitation learning and generative adversarial networks (GANs). This work derives a model-free imitation learning algorithm from this framework, which shows significant performance gains over existing model-free methods in imitating complex behaviors in large, high-dimensional environments. The methodology of GAIL involves recovering the expert's cost function with inverse reinforcement learning and then extracting a policy from that cost function with reinforcement learning. 

\textbf{TPIL} \cite{stadie2017third}\,\, This work presents a method for unsupervised third-person imitation learning (TPIL), where an agent is trained to achieve a goal by observing a demonstration from a different viewpoint. The agent receives only third-person demonstrations and does not have correspondence between teacher and student states. The primary insight of the method is to utilize domain-agnostic features obtained from recent advances in domain confusion during the training process. TPIL aims to minimize class loss while maximizing domain confusion. It addresses questions regarding the algorithm's sensitivity to hyper-parameter selection, changes in camera angle, and its performance compared to baselines. It also investigates whether the algorithm benefits from both domain confusion and velocity.

\section{Hyper-parameters}

\begin{table}[h]
\begin{center}
\begin{small}
\setlength\tabcolsep{1.2pt}
\scalebox{0.95}{\begin{tabular}{ll}
\toprule
\textbf{Hyperparameter} & \textbf{Value}  \\
\midrule
Policy hidden dim & $(128, 128)$ \\
Value hidden dim & $(128, 128)$ \\
Discriminator hidden dim &  (128, 128)\\
Embedding size & $10$ Hopper, $7$ Swimmer  \\ 
batch size & 2048\\
Optimizer & Adam \\
Learning rate & chooses from $[3\times 10^{-4}, 1\times 10^{-4}, 8\times 10^{-5}, 6\times 10^{-5}]$\\
Lambda & chooses from $[0.2, 0.5]$ \\
Frequency of update & $1$ for encoder and policy discriminator\\
&$5$, $7$ for Hopper and Swimmer noise discriminator respectively\\
Agent rate clip & $[0.01, 0.99]$\\
Confuse prob clip & [0.1, 0.9] \\
\bottomrule
\end{tabular}}
\end{small}
\end{center}
\caption{Hyperparameters of DIDA for all experiments.}
\label{tab:atari_hyperparameters}
\end{table}


\end{document}